\documentclass[acmlarge,screen,nonacm]{acmart}
\newcommand{\net}[0]{\text{PLOS-net}}
\newcommand{\cnet}[0]{\text{C-PLOS-net}}
\newcommand{\N}[0]{n}
\newcommand{\K}[0]{k}
\newcommand{\M}[0]{m}
\newcommand{\myrho}[0]{$\rho$}
\newcommand{\rmnk}[0]{\myrho\M\N\K-landscape}

\AtBeginDocument{%
  }
\usepackage{xcolor}
\usepackage{subcaption}
\setcopyright{acmlicensed}
\copyrightyear{2025}
\acmYear{2025}
\acmDOI{10.1145/3712256.3726334}
\acmConference[GECCO '25]{Genetic and Evolutionary Computation
Conference}{July 14--18, 2025}{Malaga, Spain}
\acmISBN{979-8-4007-1465-8/2025/07}

\begin{document}

\title{Customized Exploration of Landscape Features Driving Multi-Objective Combinatorial Optimization Performance}

\author{Ana Nikolikj}
\email{ana.nikolikj@ijs.si}
\orcid{1234-5678-9012}

\affiliation{%
  \institution{Jozef Stefan International
Postgraduate School}
\city{Ljubljana}
\country{Slovenia}
}
\affiliation{%
  \institution{Jozef Stefan Institute}
  \city{Ljubljana}
  \country{Slovenia}
}

\author{Gabriela Ochoa}
\email{gabriela.ochoa@stir.ac.uk}
\affiliation{%
  \institution{University of Stirling}
  \city{Scotland}
  \country{United Kingdom}
}

\author{Tome Eftimov}
\email{tome.eftimov@ijs.si}
\affiliation{%
  \institution{Jozef Stefan Institute}
  \city{Ljubljana}
  \country{Slovenia}
}

\renewcommand{\shortauthors}{Nikolikj, Ochoa, Eftimov}

\begin{abstract}
We present an analysis of landscape features for predicting the performance of multi-objective combinatorial optimization algorithms. We consider features from the recently proposed compressed Pareto Local Optimal Solutions Networks (\cnet) model of combinatorial landscapes. The benchmark instances are a set of \rmnk s with 2 and 3 objectives and various levels of ruggedness and objective correlation. We consider the performance of three algorithms -- Pareto Local Search (PLS), Global Simple EMO Optimizer (GSEMO), and Non-dominated Sorting Genetic Algorithm (NSGA-II) -- using the resolution and hypervolume metrics. Our tailored analysis reveals feature combinations that influence algorithm performance specific to certain landscapes. This study provides deeper insights into feature importance, tailored to specific \rmnk s and algorithms.
\end{abstract}




\maketitle

\section{Introduction}
\label{sec:intro}
In black-box optimization (BBO), selecting the best algorithm for a given problem is challenging due to the abundance of options. This has led to research in meta-learning for automated algorithm selection, which uses machine learning (ML) models to predict algorithm performance based on problem properties (landscape features)~\cite{liefooghe2019landscape,kostovska2023comparing}.

Extensive research in single-objective optimization (SOO) has led has led to numerous methods for extracting landscape features for continuous SOO problems, including Exploratory Landscape Analysis (ELA), Topological Landscape Analysis (TLA), and deep learning-based approaches (TransOpt, DeepELA, Doe2Vec), as summarized in a recent survey~\cite{cenikj2024survey}. These techniques are evaluated for problem classification and automated algorithm selection/configuration. Other studies compare and combine features to improve predictive models and identify truly impactful features, avoiding the introduction of less meaningful ones~\cite{cenikj2024cross,seiler2024synergies}.

Compared to SOO, multi-objective optimization (MOO) is relatively under-explored. In continuous MOO, various techniques have been studied, including cost landscapes~\cite{Fonseca1996}, gradient field heatmaps~\cite{Kerschke2017}, local dominance landscapes~\cite{Fieldsend2021}, and landscape plots illustrating optimal trade-offs~\cite{Schapermeier2020}. For combinatorial MOO, tools such as Pareto Optimal Solution Networks (PLOS-net)~\cite{Liefooghe2018} and Pareto Local Optima Networks (PLON)~\cite{Fieldsend2019} have proven valuable for analyzing optimization problem properties. More recently, funnel features~\cite{ochoa2024funnels} based on solution rankings using non-dominated sorting, along with the compressed Pareto Local Optimal Solution Network (C-PLOS-net)~\cite{liefooghe2023pareto}, have been proposed and studied to predict the performance of established local search and evolutionary algorithms. The main challenge is the lack of evaluation of feature groups to determine whether their combinations can yield improved results. Classical correlation methods or global feature importance often overlook their impact on specific problem landscapes.

\noindent\textbf{Our contribution}: 
We present a customized analysis of the importance of landscape features by tailoring it to specific \rmnk s and combinatorial MOO algorithm combination. We consider the recently proposed 10~\cnet~features in addition to the 24 \net~features from~\cite{Liefooghe2018}, and 7~\textit{funnel} metrics from~\cite{OchoaLV24}, extracted from a set of \rmnk s with 2 and 3 objectives and various levels of ruggedness and objective correlation. The combined feature groups are used to predict the performance of three algorithms: PLS~\cite{paquete2007a}, GSEMO~\cite{laumanns2004}, and NSGA-II~\cite{Deb2002}). Our study offers tailored insights into which landscape features are the most important for the performance of specific landscapes, for different MOO algorithms. Additionally, we identify landscape feature combinations that make certain \rmnk s particularly hard for some algorithms. The analysis is conducted twice, using two distinct performance metrics: resolution and hypervolume. In summary, unlike previous studies that evaluate global feature importance across all algorithms and different \rmnk s, this analysis is tailored to specific \rmnk s, algorithm and performance metric combination. This analysis offers deeper insights into the relevance of various feature groups and helps identify features that make certain landscapes particularly challenging for specific algorithms to solve. The results show that the number of interactions is the most influential benchmark parameter affecting the hardness of problem instances for the algorithms. It also shows that combining two feature sets is crucial, as they complement each other and are key for analyzing specific landscapes with different algorithms.

\noindent\textbf{Outline}: Section~\ref{sec:realted_work} reviews the background on combinatorial MOO and summarizes previous landscape analyses. Section~\ref{sec:methodology} briefly revisits the methodology employed for the analysis. Section~\ref{sec:exp_design} outlines the experimental design and data used. In Section~\ref{sec:discussion}, we discuss the strengths and limitations of the analysis. Finally, Section~\ref{sec:conclusions} concludes the paper and offers directions for future research.

\noindent\textbf{Data and code availability:} The data and code used in this study are available at~\cite{Nikolikj2025zenodo}.

\vspace{-1em}
\section{Background}
\label{sec:realted_work}
Here, we briefly explain combinatorial MOO and the $\rho$mnk-landscape benchmark suite. Then, we provide an overview of related work on landscape analysis for the benchmark suite.

\vspace{-1em}
\subsection{Combinatorial multi-objective optimization}
Combinatorial MOO involves finding the optimal solution to optimization problems characterized by decision variables that take on discrete values and multiple conflicting objectives~\cite{coello2010multi}. Solving MOO problems requires identifying trade-offs among objectives and determining a set of non-dominated solutions (i.e. Pareto optimal solutions). The collection of Pareto optimal solutions is known as the Pareto set, and its representation in the objective space is referred to as the Pareto front. MOO algorithms aim to identify or provide a good approximation of the Pareto set/front.

\vspace{-1em}
\subsection{\texorpdfstring{$\rho$MNK Landscapes}{rhoMNK Landscapes}}
\rmnk s~\cite{Verel2013} is a benchmark suite for combinatorial MOO that consists of multi-objective multi-modal combinatorial problems with objective correlation, extending the the benchmark suite of single-objective and nk-landscapes with independent objectives~\cite{Kauffman1993, Aguirre2007, knowles2007}. The solution space is $X = \lbrace 0, 1 \rbrace^{n}$, where $n$ denotes the problem size or the total number of binary decision variables. The objective vector $f=(f_1, \ldots, f_i, \ldots, f_m)$ maps $f\colon \lbrace 0, 1 \rbrace^{n} \to [0,1]^m$, where $m$ is the number of objectives and each objective $f_i$ is maximized. The value $f_i(x)$ for a solution $x=(x_1, \ldots, x_j, \ldots, x_n)$ is the average of the contributions from each variable~$x_j$. The contribution of~$x_j$ depends on its value and the values of $k < n$ other variables, selected uniformly at random from the $(n-1)$ variables excluding~$x_j$\cite{Kauffman1993}. By adjusting $k$, the problem landscape transitions from smooth to rugged. In \rmnk s, the contribution values follow a multivariate uniform distribution, with the correlation coefficient $\rho > \frac{-1}{m-1}$ controlling the objectives' conflict level~\cite{Verel2013}. Positive $\rho$ reduces conflict, while negative $\rho$ increases it. Notably, in the \rmnk s suite poses varying challenges for multi-objective algorithms~\cite{Daolio2017,Liefooghe2020}.

\vspace{-1em}
\subsection{Landscape analysis}
The study of MOO landscapes is relatively limited compared to the extensive literature on SOO landscapes. In the case of combinatorial MOO, several studies address the landscape analysis of MOO combinatorial $\rho$mnk landscapes.

In~\cite{liefooghe2023pareto}, the authors extend the concept of the PLOS-net by proposing the C-PLOS-net. Statistical metrics are calculated for both networks as mathematical objects and to predict algorithm performance with a Random Forest (RF)~\cite{biau2016random} and algorithm selection via a Decision Tree (DT). Global importance derived from the RF model reveals the relevance of the features as descriptors of diverse problem landscapes. Additionally, the Pearson coefficient and Spearman's rank showed that most features were significantly correlated to the performance of the analyzed algorithms. A total of 24 features can be computed for the PLOS-net, while the C-PLOS-net allows for the calculation of 10 additional ones. Recently, five new funnel features have been introduced to characterize the presence of funnels in $\rho$mnk landscapes. These features are derived from solution ranks using non-dominated sorting -- and a variation of the graph-based model for MOO landscapes, the C-PLOS-net. The funnel features, combined with the benchmark parameters $\rho$, $m$, $n$, and $k$, have been utilized to predict algorithm performance using a Random Forest model and to perform algorithm selection via a Decision Tree. The proposed funnel features effectively capture the landscape’s global structure, exhibit strong correlations with the benchmark parameters, and provide valuable insights into the performance of established multi-objective local search and evolutionary algorithms. The mentioned landscape feature groups, the 24 base PLOS-net features, the additional 10 C-PLOS-net, and the five funnel features have been studied individually, but a comprehensive analysis combining them for algorithm performance prediction is still lacking. Additionally, only the global importance of these features has been assessed, although some features may be more relevant for a specific $\rho$mnk landscape-algorithm combination. 

In~\cite{liefooghe2019landscape}, dominance and (hypervolume) metric-based landscape features have been proposed. A subsequent study~\cite{liefooghe2020dominance} demonstrated that these features correlate with the performance of dominance-based MOEAs, but their correlation with the performance of decomposition MOEAs was found to be less significant. This observation motivated the development of decomposition-based landscape features in~\cite{cosson2021decomposition}. The features are used to predict the $\rho$ and $k$ benchmark parameters of different $\rho$mnk landscapes and to support algorithm selection among three variants of the Multi-Objective Evolutionary Algorithm based on Decomposition (MOEA/D)~\cite{zhang2007moea}, leveraging Random Forest models for these tasks. All the proposed sets of features are interesting and complementary. One advantage of C-PLOS-net features is that they  are independent of any particular algorithm or performance metric, at least when they are based on a full enumeration of the search space as is the case of the instances studied in this article.

\vspace{-1em}
\section{Methodology}
\label{sec:methodology}
We used the recently proposed \textit{benchmarking algorithm footprints} approach~\cite{nikolikj2024comparing} to analyze similarities and differences among SOO algorithms in BBO, following these steps:

\noindent\textbf{(1) Meta-Learning for Performance Prediction}: This approach involves training a multi-target regression (MTR) model to simultaneously predict the performance of multiple algorithms based on the landscape features of a problem instance. Each problem instance has a single feature representation, capturing its inherent landscape properties. The MTR model leverages the same feature representation to predict the performance of multiple algorithms~\cite{zhen2017multi}. 

\noindent\textbf{(2) Meta-Representation Creation}: This step involves applying an explainability method to the trained MTR model to compute~\textit{local feature importance}, resulting in feature importance scores for each problem instance in the dataset. Given that MTR models predict the performance of multiple algorithms, the local feature importance scores are computed for each algorithm separately. 
The local feature importance scores serve as the basis for constructing~\textit{meta-representations}, that show which landscape feature combination is important for the performance of the algorithm, capturing the relationship between the landscape properties of the problem instance and the algorithm behavior. Performing this for all algorithms, results in tailored meta-representations to each problem instance-algorithm pair.

\noindent\textbf{(3) Clustering of Meta-Representations}: A clustering algorithm is used to automatically group the meta-representations, and identify clusters of problem instance-algorithm pairs where a similar landscape feature combination is important, suggesting similar algorithm behavior, and different algorithm behavior between clusters.

\noindent\textbf{(4) Footprint Analysis and Comparison}: By examining the distribution of meta-representations associated with an algorithm across the clusters, we define its unique~\textit{algorithm footprint}. By benchmarking the footprints across algorithms, we reveal performance similarities, differences, and easy and hard landscapes for the algorithm portfolio as a whole.

\noindent\textbf{(5) Important Feature Identification}: Identify which landscape feature combination is important in each cluster, by aggregating the feature importance score over the meta-representations in the cluster. Local feature importance scores can be analyzed at multiple levels: for a single problem instance to compare algorithms or at the cluster level by aggregating their scores, highlighting landscape features that impact algorithms performance.

\vspace{-1em}
\section{Experimental design}
\label{sec:exp_design}
\noindent\textbf{Benchmark suite}: In the $\rho$mnk-landscapes benchmark suite different problem instances can be generated by varying the benchmark parameters $\rho$, $m$, $n$, and $k$. Table~\ref{tab:param_problems} presents the ranges for the benchmark parameters, with a total of 18 combinations. For each benchmark parameter combination, 10 problem instances are randomly generated, resulting in a total of 180 $\rho$mnk problem instances. The parameter choices enable the analysis of landscapes ranging from smooth to rugged, with two or three objectives that may be conflicting, uncorrelated, or correlated.

\begin{table}
\caption{Benchmark parameters for generating $\rho$mnk problem instances, with 10 random instances per parameter combination.}
\footnotesize
\centering
\begin{tabular}{l|l}
\toprule
\textbf{description}    &   ~\textbf{values}              \\
\midrule
number of variables		&	~$n = 16$                    \\
number of interactions~ &	~$k \in \{1, 2, 4\}$    \\
number of objectives    &	~$m \in \{2, 3\}$         \\
objectives correlation	&	~$\rho \in \{-0.4,  0.0, 0.4\}$   \\
\bottomrule
\end{tabular}
\label{tab:param_problems}
\end{table}

\noindent\textbf{Landscape features}: The landscape characteristics of $\rho$mnk problem instances are described using 41 features: 24 base~\net~features, 10 \cnet~features, and seven funnel features. The funnel features include five from~\cite{OchoaLV24} and two new ones:~\textsf{pos\_num} (number of Pareto optimal solutions) and~\textsf{pos\_strength} (the incoming weighted degree to Pareto nodes).

\noindent\textbf{Algorithm portfolio}: We used three widely used multi-objective algorithms for solving the $\rho$mnk problem instances: PLS, a local search method that maintains an unbounded archive of mutually non-dominated solutions;  GSEMO, a global elitist steady-state evolutionary algorithm; and NSGA-II, a dominance-based evolutionary algorithm.

\noindent\textbf{Performance data}: Performance data is taken from a prior study\\~\cite{liefooghe2023pareto}, with each algorithm run 30 times per problem instance. PLS is terminated upon reaching a Pareto local optimum set~\cite{paquete2007a}, while G-SEMO and NSGA-II are stopped after 10,000 evaluations. NSGA-II uses a population size of 100 (further presented on the plots as~\textsf{nsga\_pop}). Algorithm performance was evaluated using~\emph{pareto resolution} (\textsf{reso}), representing the proportion of identified Pareto optimal solutions, and~\emph{relative hypervolume} (\textsf{hv}), which measures the hypervolume of the final archive relative to the exact Pareto front, the reference point was set to the origin). Higher values indicate better performance for both metrics. 

\noindent\textbf{MTR predictive model}: To find a good MTR model, we conduct experiments with three machine learning algorithms: Random Forest (RF)~\cite{biau2016random}, Neural Network (NN)~\cite{glorot2010understanding}, and Multi-Task Elastic Net (MTEN)~\cite{zou2005regularization}. The RF and MTEN models are implemented using the~\textit{scikit-learn} library~\cite{pedregosa2011scikit}, while the NN model is built using the \textit{keras} framework~\citep{chollet2015keras}, in Python. As a data prepossessing step, each feature has been normalized in the range 0 to 1, to bring the features to the same scale as this can affect the prediction task. The scaling parameters are fitted on the training dataset, and then the test dataset is transformed using the same parameters. 

\noindent\textbf{MTR model evaluation}: 
The dataset for training the MTR models includes 180 problem instances, with 10 problem instances generated for each of the 18 benchmark parameter combinations. A dataset split allocates one instance per benchmark parameter combination for testing (18 problem instances) and the remaining 162 problem instances for training. The model performance is evaluated using 9-fold stratified cross-validation on the training set, which allocates one instance per benchmark parameter combination for validation. Models are trained nine times, with eight folds for training and one for validation, and the average validation performance determines the best model. The final model is retrained on the full training set and evaluated on the test set using Mean Absolute Error (MAE) and R-squared (R²). MAE measures average absolute prediction error, while R² quantifies the proportion of variance in the ground-truth labels that is explained by the model's predictions.

\noindent\textbf{MTR model hyper-parameter tuning}: Optimal hyperparameters for the MTR models are determined using the Tree-structured Parzen Estimator (TPE) algorithm from the Optuna library~\cite{akiba2019optuna}. The search, with a budget of 50 trials, begins with 40 Random Search trials to broadly explore the space, followed by 10 TPE trials to focus on promising regions. The hyperparameters yielding the best cross-validation score on the training set are selected.

\noindent\textbf{Feature selection}: After hyperparameter tuning, we applied Sequential Forward Feature Selection (SFFS)~\citep{ferri1994comparative} to potentially further improve MTR model performance. Starting with an empty feature set, SFFS iteratively adds the feature that improves the performance the most. This process continues until all features are added. We applied SFFS to the RF and MTEL models but omitted it for the NN model, as due to high computational cost it is not common practice. The feature subset with the best cross-validation score was selected.

\begin{figure}[t]
    \begin{subfigure}[b]{0.3\textwidth}
        \centering
        \includegraphics[width=\linewidth]{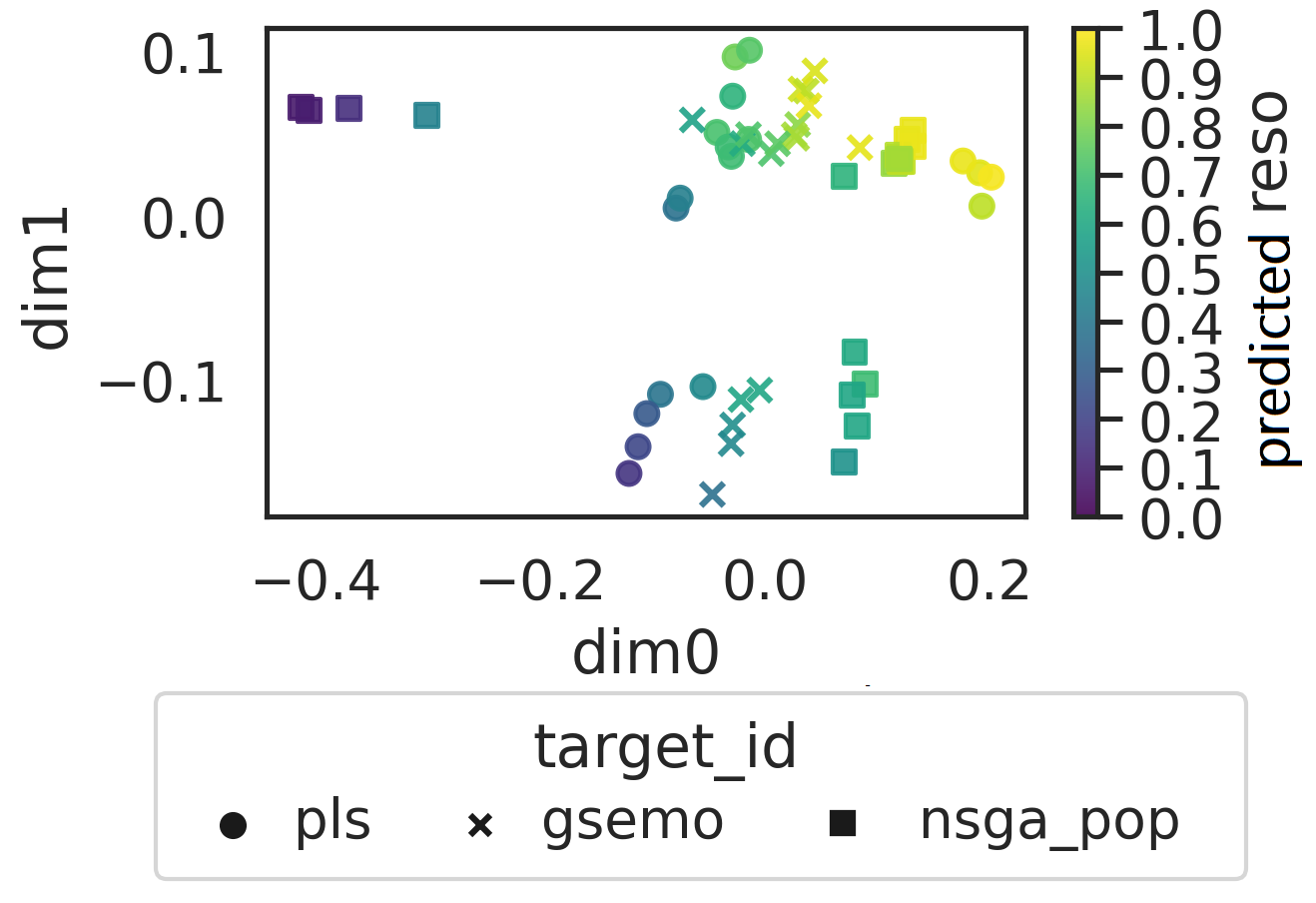}
        \caption{}
        \label{fig:predicted_2d_reso}
    \end{subfigure}
    \hspace{3cm} 
    \begin{subfigure}[b]{0.40\textwidth}
        \centering
        \includegraphics[width=\linewidth]{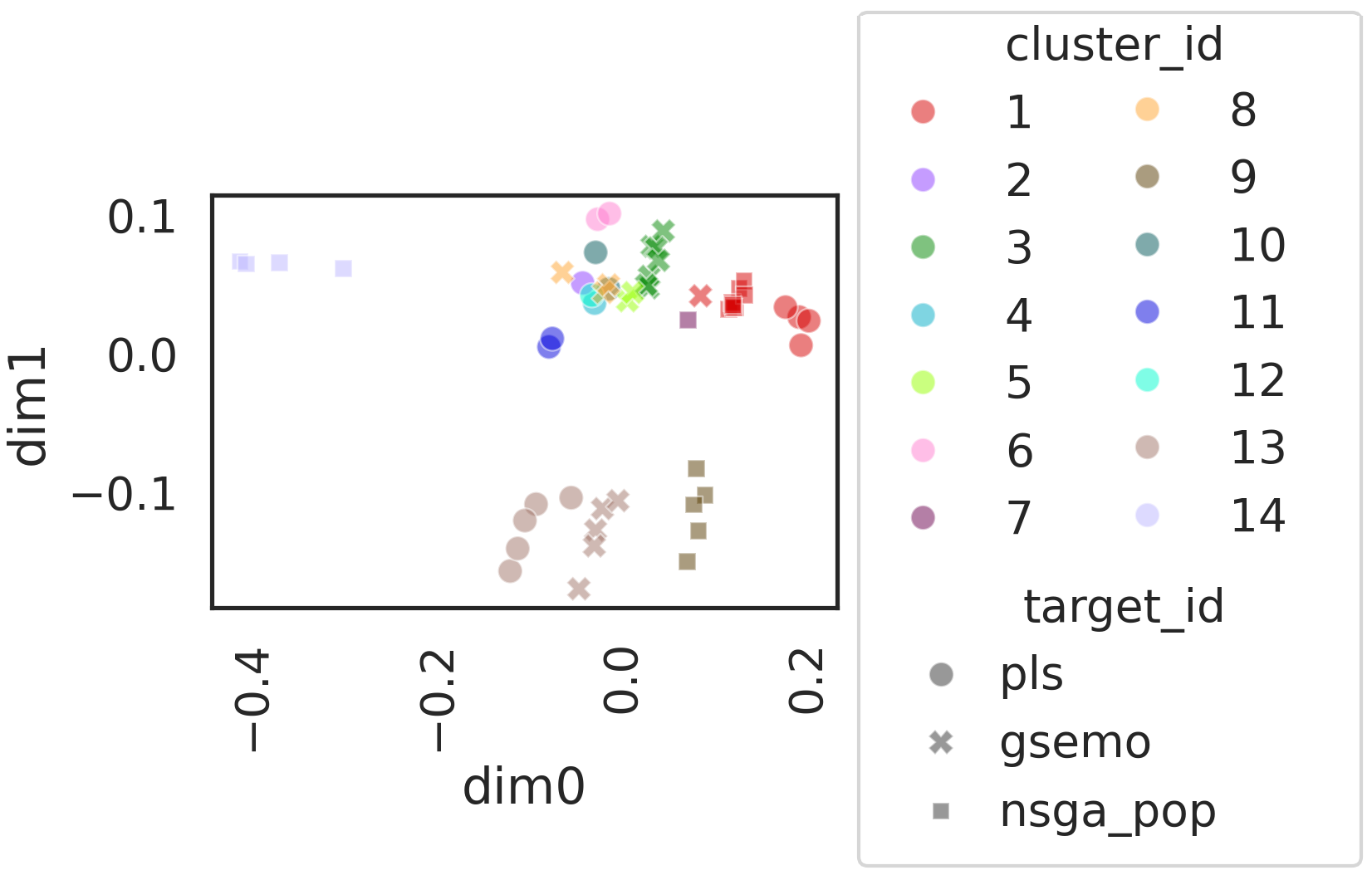}
        \caption{}
        \label{fig:clusters_2d_reso}
    \end{subfigure}
    \caption{2D visualization of $\rho$mnk landscapes using SHAP meta-representation from the test data, showing (a) ground-truth \textsf{reso} performance for three algorithms and (b) clustering results. Axes dim0 and dim1 represent the projected 2D dimensions.}
    \Description{}
    \label{fig:shap_2d}
\end{figure}

\noindent\textbf{Model Selection}: We compared the performance of the MTR models to a baseline which always predicts the average algorithm performance as calculated on the training set. The RF model with hyperparameter tuning and feature selection achieved the lowest cross-validation error and is used for the subsequent analysis. The selected hyperparameters area: using all available features to build the trees, the maximal depth of the trees of seven, 192 trees in the forest, minimum samples to split of three, and minimum samples per leaf of one. The feature selection process identified 23 features as the optimal subset for the RF model.

\noindent\textbf{SHAP local feature importance}: 
Landscape features describe $\rho$mnk landscapes without directly linking to algorithm performance. Using the MTR model with SHAP~\cite{van2022tractability}, we calculate local feature importance to show how features influence performance predictions. SHAP values link features to performance via $q$-dimensional meta-representations $s_1, s_2, \ldots, s_q$, where $q$ is the number of features. Each $\rho$mnk instance has multiple SHAP meta-representations (e.g., one per algorithm), enabling comparison of feature patterns driving algorithm performance. Calculations use the SHAP library~\cite{NIPS2017_7062} in Python.

\noindent\textbf{Clustering}: We use hierarchical clustering~\cite{tokuda2022revisiting} to automatically group the SHAP meta-representations of the problem instances. To assess the quality of the clusters, we use the Silhouette coefficient, which ranges from 0 to 1 and evaluates how distinct and compact the clusters are. A higher Silhouette score reflects better-defined clusters. We optimize the hyperparameters of the hierarchical clustering algorithm and choose the configuration that yields the highest Silhouette score. This resulted with the highest silhouette score of 0.63, 14 as an optimal number of clusters, $cosine$ as distance metric and $average$ linking.

\section{Results}
\label{sec:results}
We present a visualization of algorithm behavior on the $\rho$mnk problem instances, along with the footprints for each algorithm. Finally, we identify the key landscape features and their combination, tailoring our analysis to each benchmark parameter combination and to the identified performance regions of the footprints.

\begin{figure}[thb]
    \centering
    \begin{subfigure}[t]{0.32\textwidth}
        \centering
        \includegraphics[width=\textwidth]{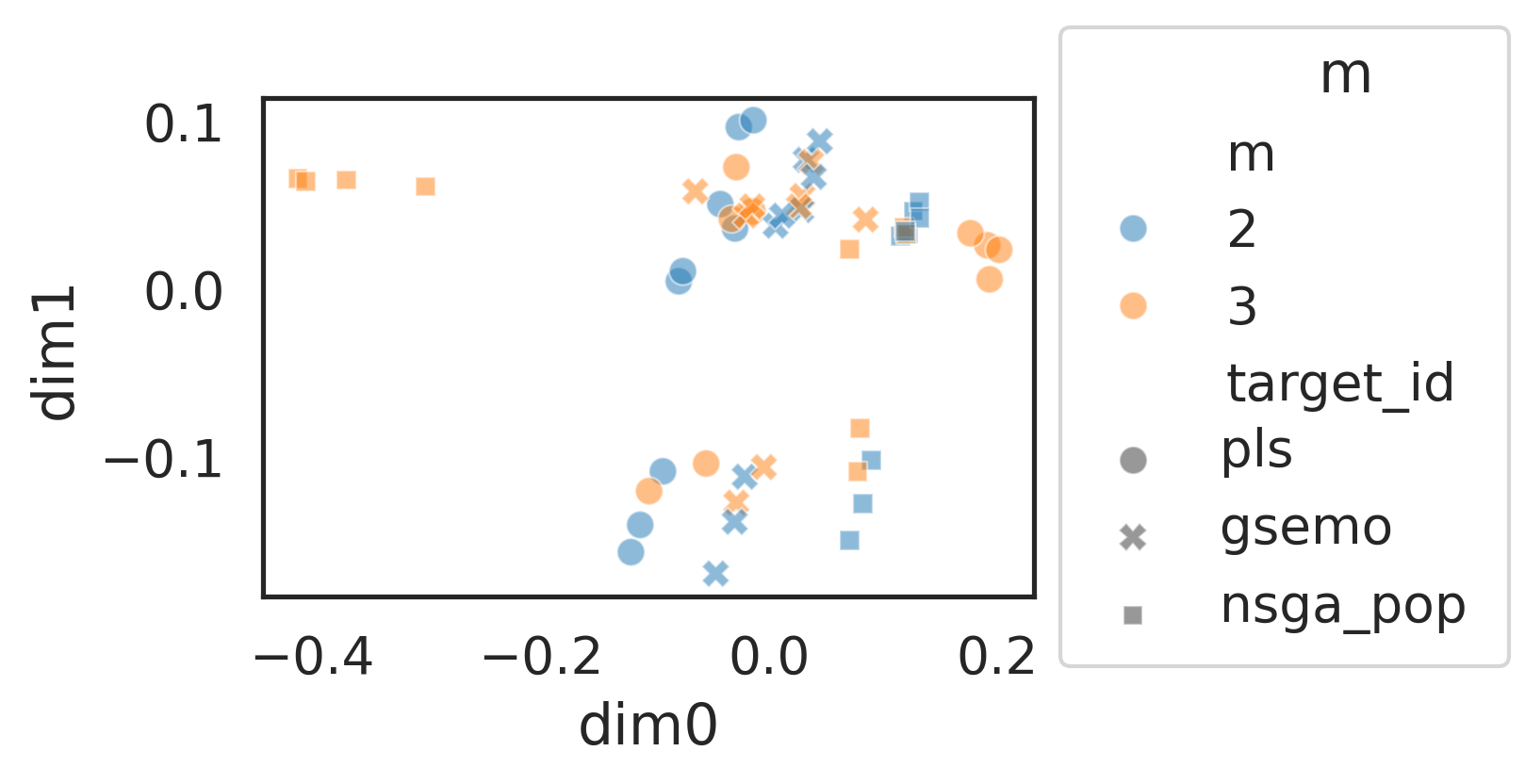}
        \caption{Number of objectives.}
        \label{fig:num_obj}
    \end{subfigure}
    \begin{subfigure}[t]{0.32\textwidth}
        \centering
        \includegraphics[width=\textwidth]{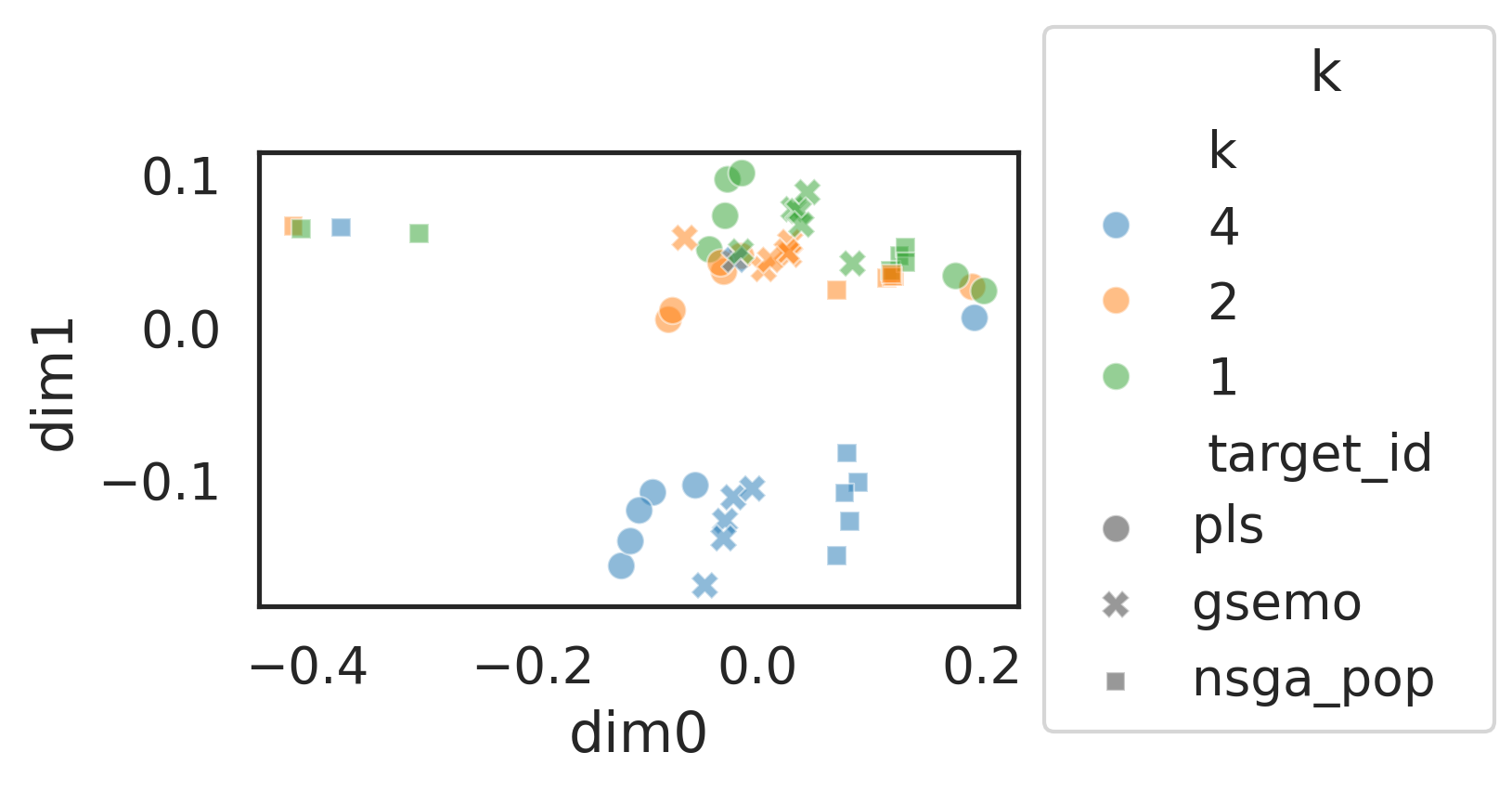}
        \caption{Number of interactions.}
        \label{fig:num_inter}
    \end{subfigure}
    \begin{subfigure}[t]{0.32\textwidth}
        \centering
        \includegraphics[width=\textwidth]{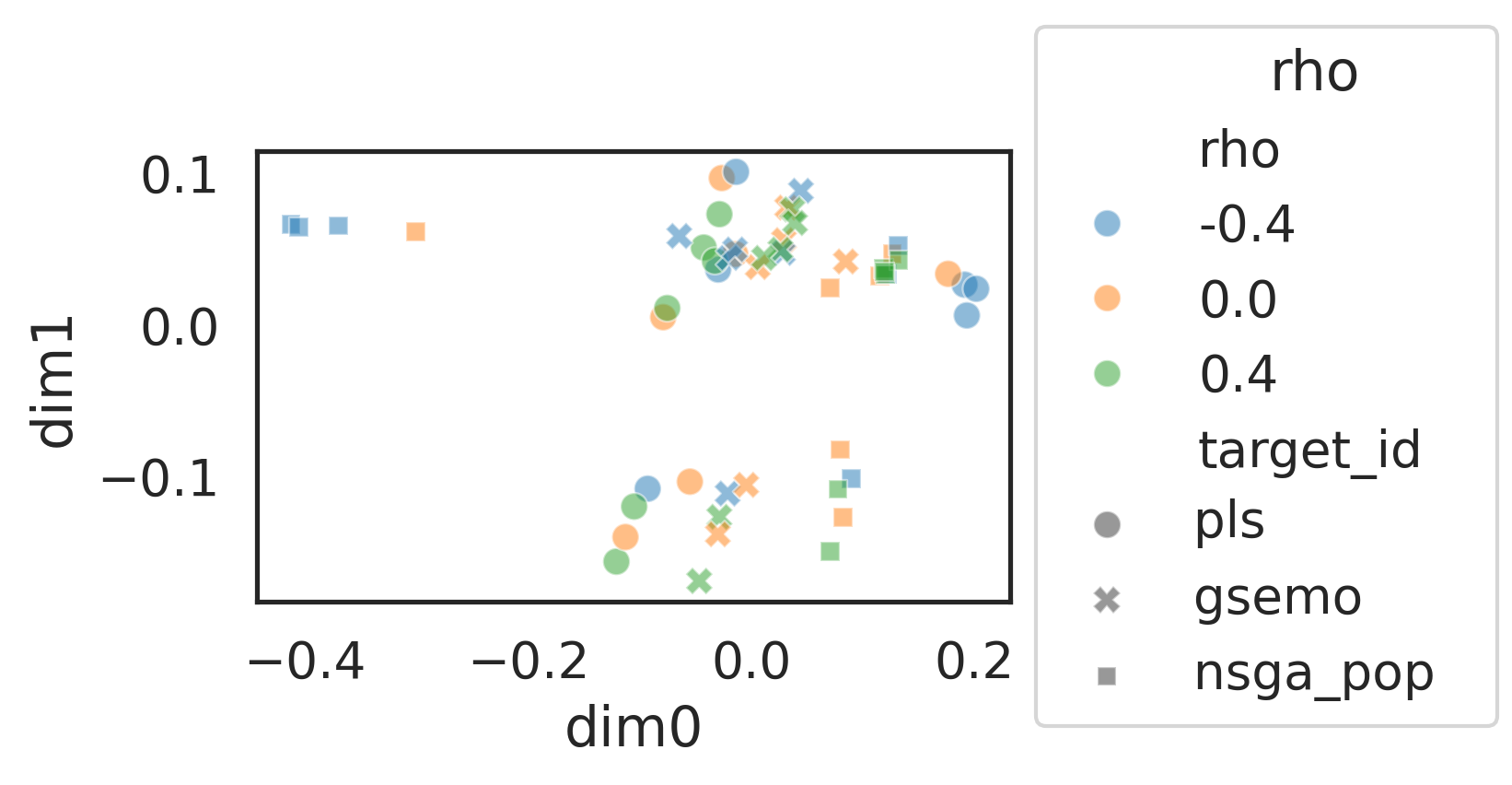}
        \caption{Objective correlation.}
        \label{fig:obj_corr}
    \end{subfigure}
    \caption{2D visualization of $\rho$mnk landscapes using SHAP meta-representation from the test data, showing the benchmark parameters.}
    \Description{}
    \label{fig:bench_parameters_2d}
\end{figure}

\subsection{Algorithm footprints}
\subsubsection{Resolution (\textsf{reso}) performance metric}
Figure~\ref{fig:shap_2d} visualizes the meta-representations corresponding to the pairings of the 18 problem instances from the test dataset with the three algorithms. Principal Component Analysis (PCA)~\citep{jolliffe2016principal} is employed as a dimensionality reduction technique to map the high-dimensional meta-representations into two dimensions for visualization. The distinct marker shapes distinguish between meta-representations associated with different algorithms, as detailed in the legend. In Figure~\ref{fig:predicted_2d_reso} the color coding reflects the predicted~\textsf{reso} performance achieved on the problem instances, where lower values indicate poor performance (dark blue) and higher values indicate good performance (yellow). The $reso$ metric measures the proportion of identified Pareto optimal solutions. The plot reveals that meta-representations that correspond to problem instances on which algorithms have poor performance are positioned towards the bottom-left side of the plot, the ones with medium algorithm performance are scattered around the center of the plot, while those corresponding to problem instances on which the algorithms have good performance are located towards the top-right corner. This indicates that based on the meta-representations (i.e. important feature combinations) we can effectively distinguish between different algorithm behaviors. 
\begin{figure}[thb]
    \centering
    \includegraphics[width=0.45\textwidth]{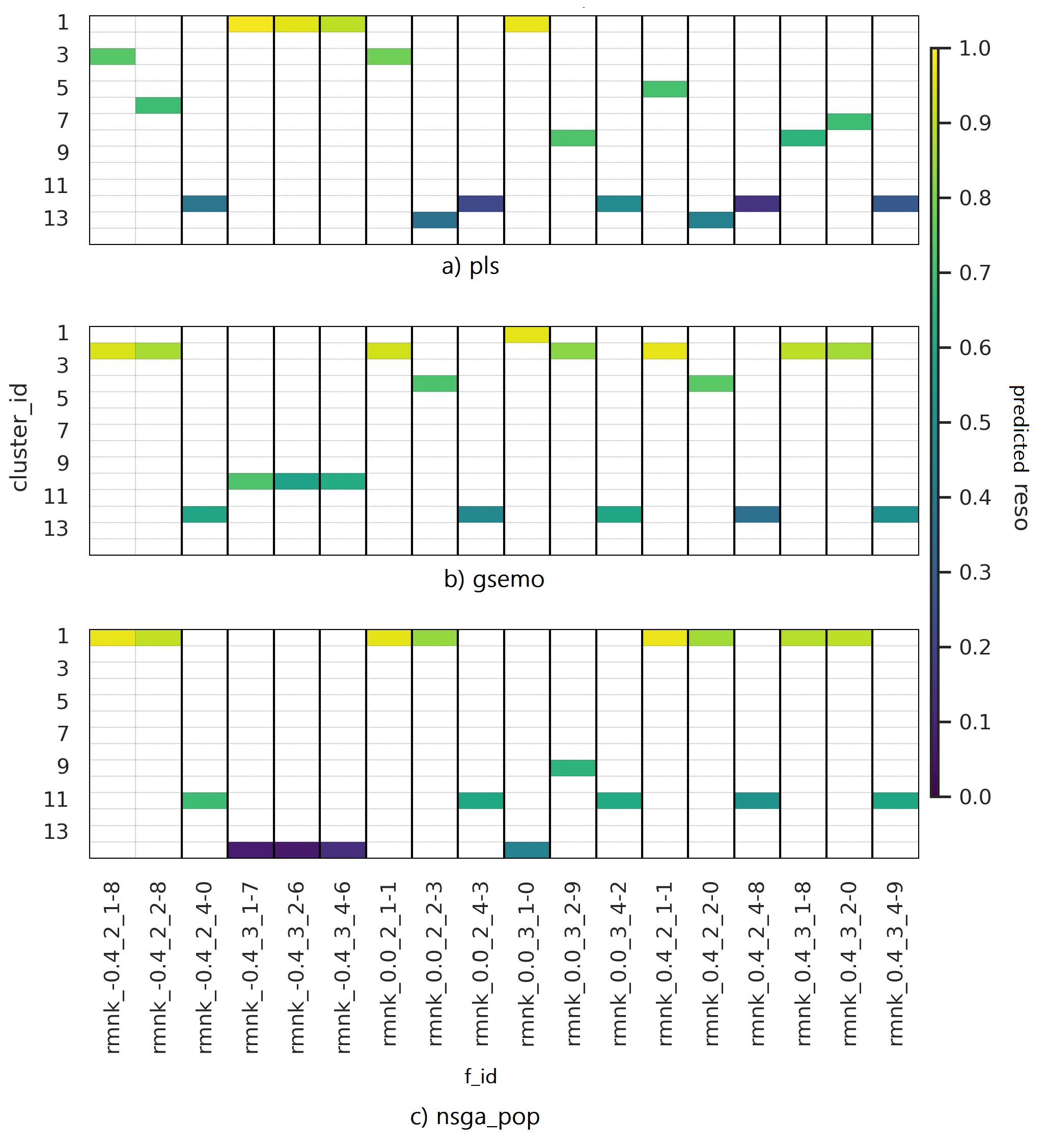}
    \caption{Algorithm footprints shown as contingency matrices: the y-axis represents 14 performance regions, and the x-axis denotes $\rho$mnk landscapes with varying benchmark parameters. Subfigures show footprints for PLS (top), GSEMO (middle), and NSGA-II (bottom).}
    \Description{}
    \label{fig:footprint_reso}
\end{figure}


\begin{figure*}[th]
    \centering
    \begin{subfigure}[t]{0.25\textwidth}
        \centering
        \includegraphics[width=\textwidth]{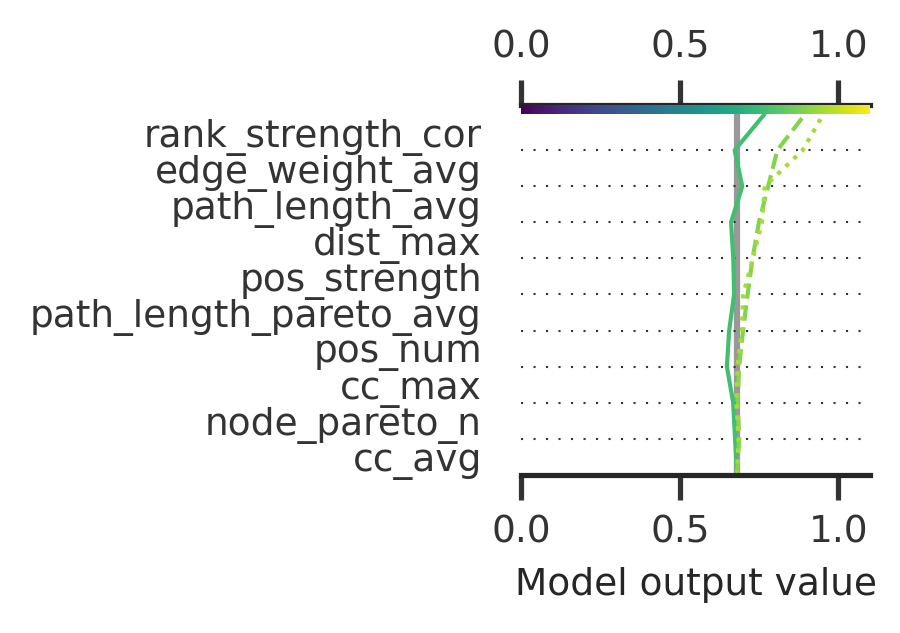}
        \caption{$\rho=-0.4$, $m=2$, $k=1$}
        \label{fig:subfig1}
    \end{subfigure}
    \hfill
    \begin{subfigure}[t]{0.23\textwidth}
        \centering
        \includegraphics[width=\textwidth]{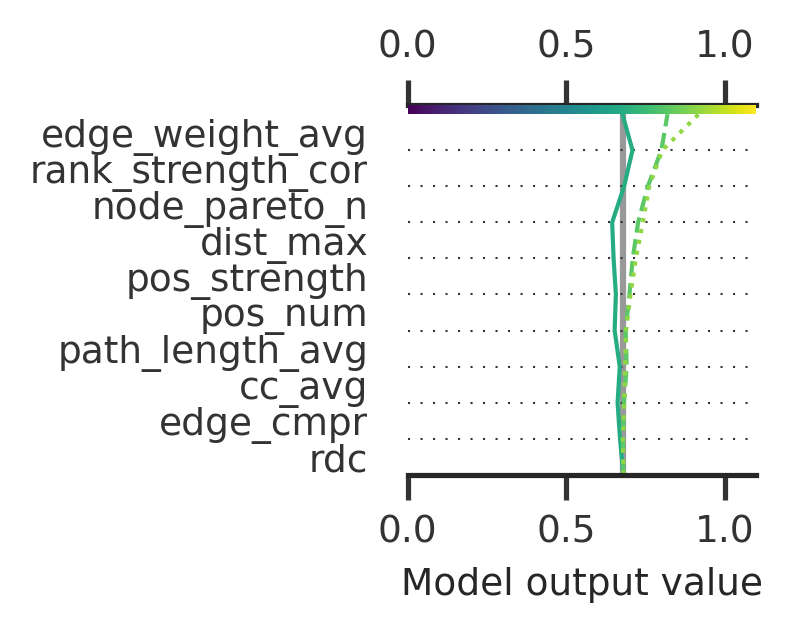}
        \caption{$\rho=-0.4$, $m=2$, $k=2$}
        \label{fig:subfig2}
    \end{subfigure}
    \hfill
    \begin{subfigure}[t]{0.25\textwidth}
        \centering
        \includegraphics[width=\textwidth]{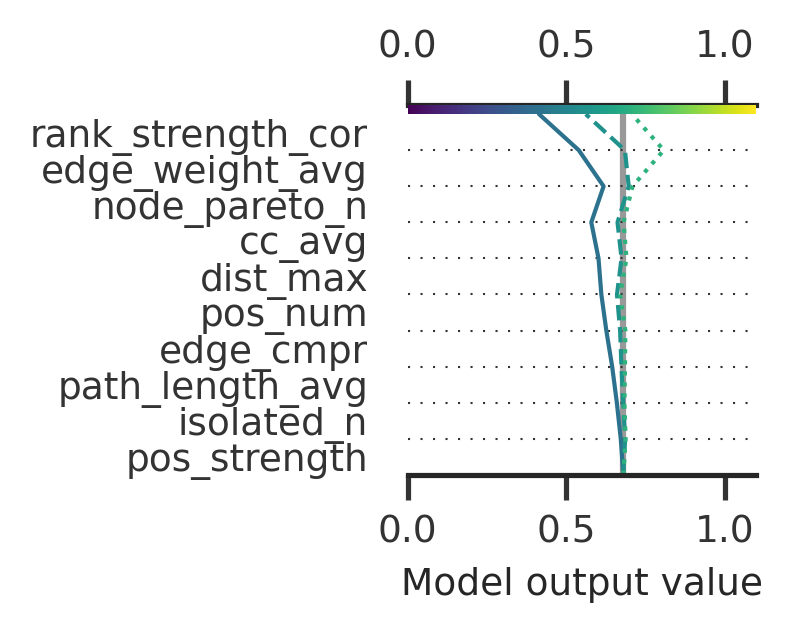}
        \caption{$\rho=-0.4$, $m=2$, $k=4$}
        \label{fig:subfig3}
    \end{subfigure}
    
    \vspace{0.1cm} 
    
    \begin{subfigure}[t]{0.23\textwidth}
        \centering
        \includegraphics[width=\textwidth]{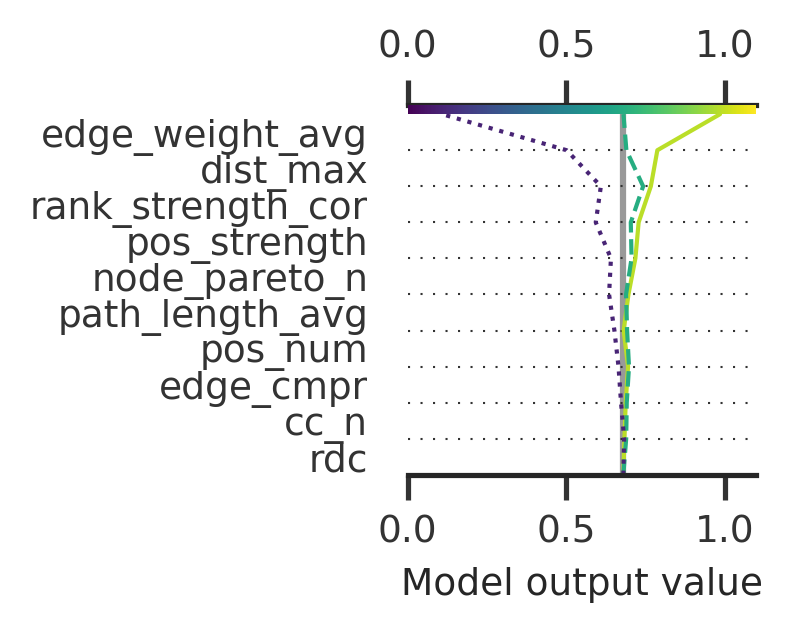}
        \caption{$\rho=-0.4$, $m=3$, $k=1$}
        \label{fig:subfig4}
    \end{subfigure}
    \hfill
    \begin{subfigure}[t]{0.25\textwidth}
        \centering
        \includegraphics[width=\textwidth]{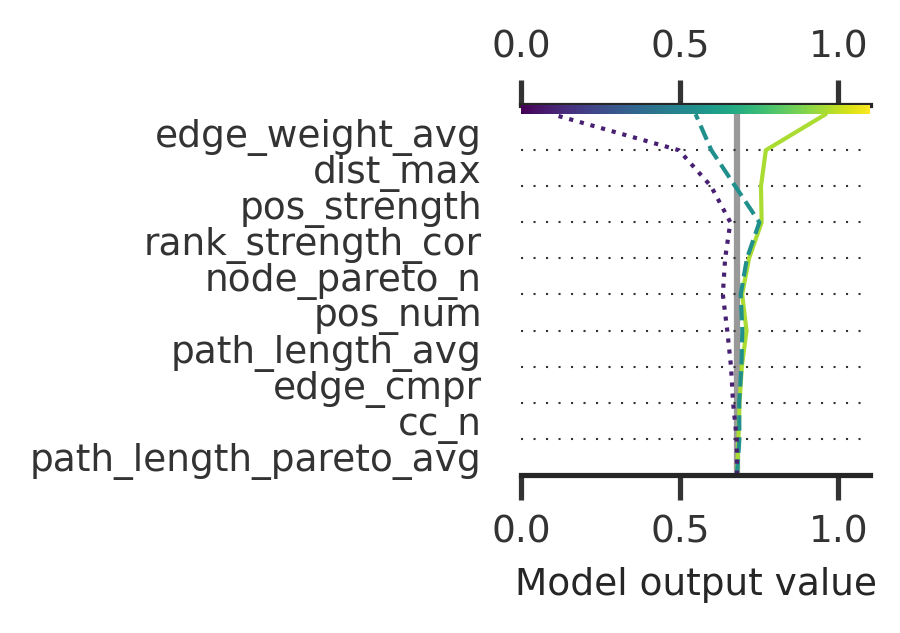}
        \caption{$\rho=-0.4$, $m=3$, $k=2$}
        \label{fig:subfig5}
    \end{subfigure}
    \hfill
    \begin{subfigure}[t]{0.25\textwidth}
        \centering
        \includegraphics[width=\textwidth]{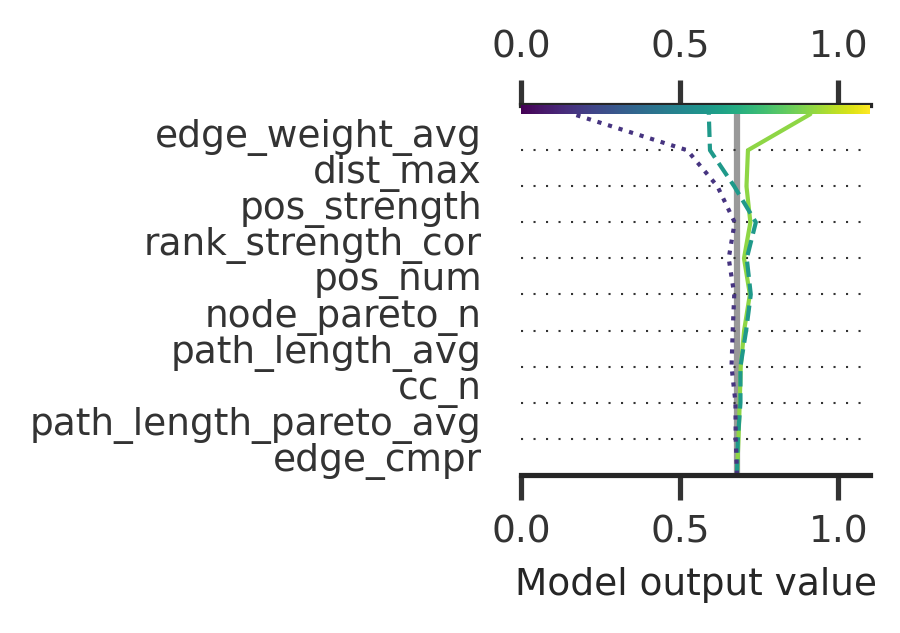}
        \caption{$\rho=-0.4$, $m=3$, $k=4$}
        \label{fig:subfig6}
    \end{subfigure}
    
    \vspace{0.1cm} 
    \raisebox{23pt}{ 
    \begin{subfigure}[t]{0.25\textwidth}
        \centering
        \includegraphics[width=\textwidth]{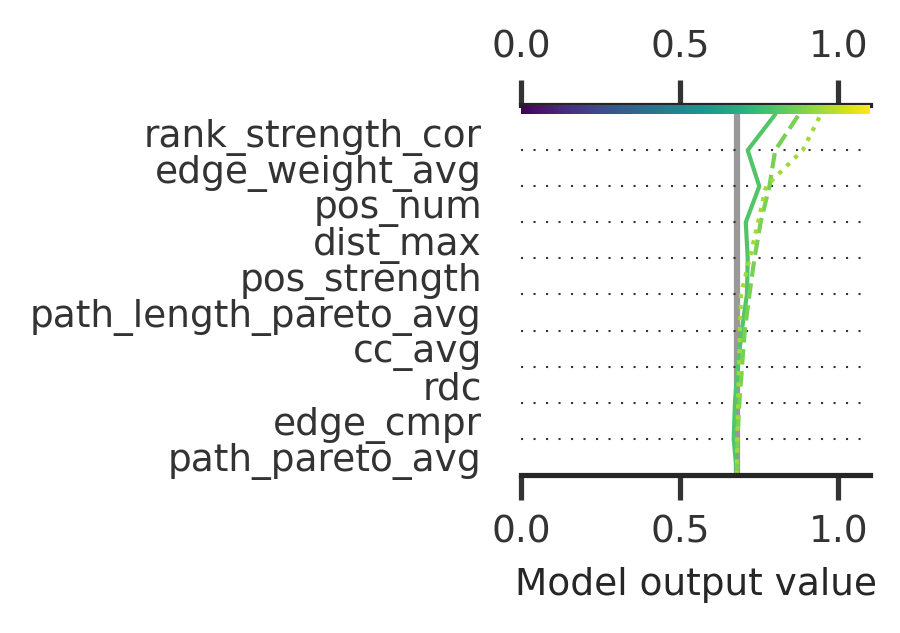}
        \caption{$\rho=0.0$, $m=2$, $k=1$}
        \label{fig:subfig7}
    \end{subfigure}}
    \hfill
    \begin{subfigure}[t]{0.3\textwidth}
        \centering
        \includegraphics[width=\textwidth]{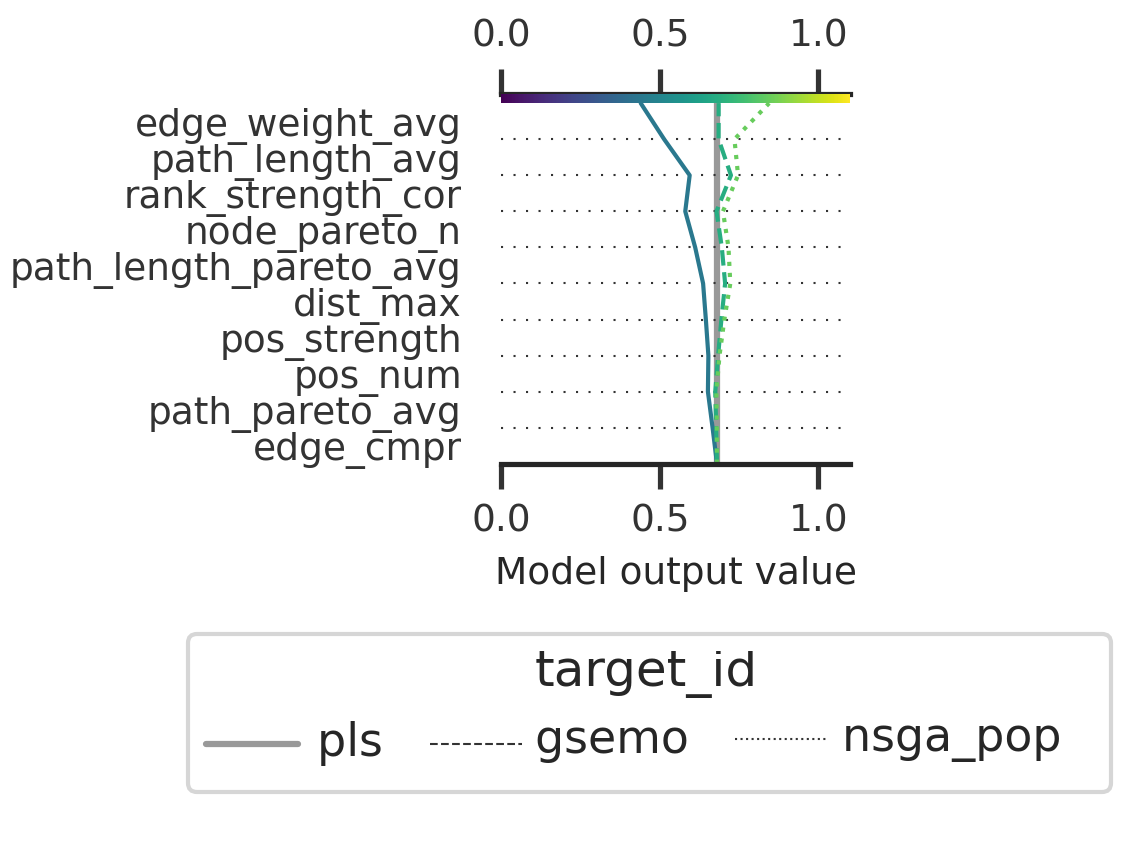}
        \caption{$\rho=0.0$, $m=2$, $k=2$}
        \label{fig:subfig8}
    \end{subfigure}
    \hfill
     \raisebox{23pt}{ 
    \begin{subfigure}[t]{0.25\textwidth}
        \centering
        \includegraphics[width=\textwidth]{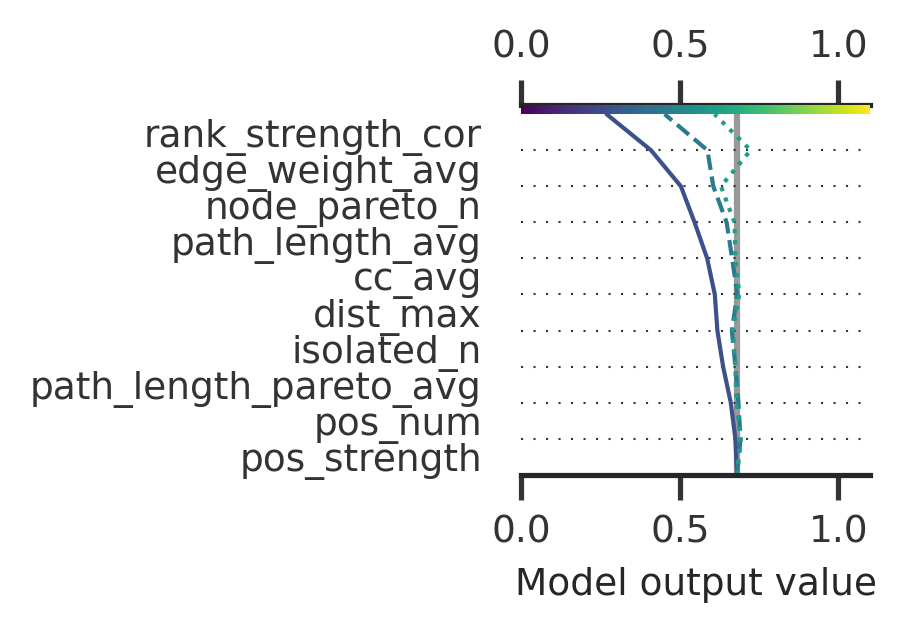}
        \caption{$\rho=0.0$, $m=2$, $k=4$}
        \label{fig:subfig9}
    \end{subfigure}}
    \caption{Key landscape features for $\rho$mnk landscapes are shown, with each plot representing a landscape and three lines for algorithm-feature interactions influencing performance: $\rho \in \{-0.4, 0\}$, $m \in \{2,3\}$, $k \in \{1,2,4\}$. The x-axis shows the predicted $reso$ performance.} 
    \Description{}
    \label{fig:post_hoc_first}
\end{figure*}

While the plot reveals roughly three performance groups, we further cluster the meta-representations for automatic grouping. Figure~\ref{fig:clusters_2d_reso} visualizes the clustering of the meta-representations, which resulted into 14 clusters. This suggests the existence of 14 distinct important landscape feature combinations. Combining the two figures, we can see how the position of the clusters corresponds to the predicted~\textsf{reso} performance. We enumerated the clusters by average predicted~\textsf{reso} performance in the cluster, in decreasing order. The first cluster contains meta-representations with good~\textsf{reso} performance on average, while the last 14th cluster contains meta-representations with poor~\textsf{reso} performance on average. It is important to note that while some clusters may exhibit similar~\textsf{reso} performance (e.g. clusters with IDs one, two, and three all have good performance), different landscape feature combinations are important in those clusters, indicating that both combinations of landscape features can be responsible for predicting similar~\textsf{reso} performance.

To examine how the meta-representations correspond to the benchmark parameters used to generate the $\rho$mnk problem instances, the color coding in Figure~\ref{fig:bench_parameters_2d} reflects the benchmark parameters: number of objectives $m$, number of interactions $k$, and the objective correlation degree $\rho$. The results reveal a clear pattern - good~\textsf{reso} performance (top-right corner of the visualizations) corresponds to problem instances with fewer interactions ($k=1$), while bad~\textsf{reso} performance corresponds with the increase in the number of interactions($k=4$). The other two benchmark parameters do not exhibit a clear pattern in relation to the performance. This suggests that the $k$ benchmark parameter has the most influence on the hardness of the problem instances for the three algorithms.
\begin{figure*}[ht]
    \centering
    \begin{subfigure}[t]{0.25\textwidth}
        \centering
        \includegraphics[width=\textwidth]{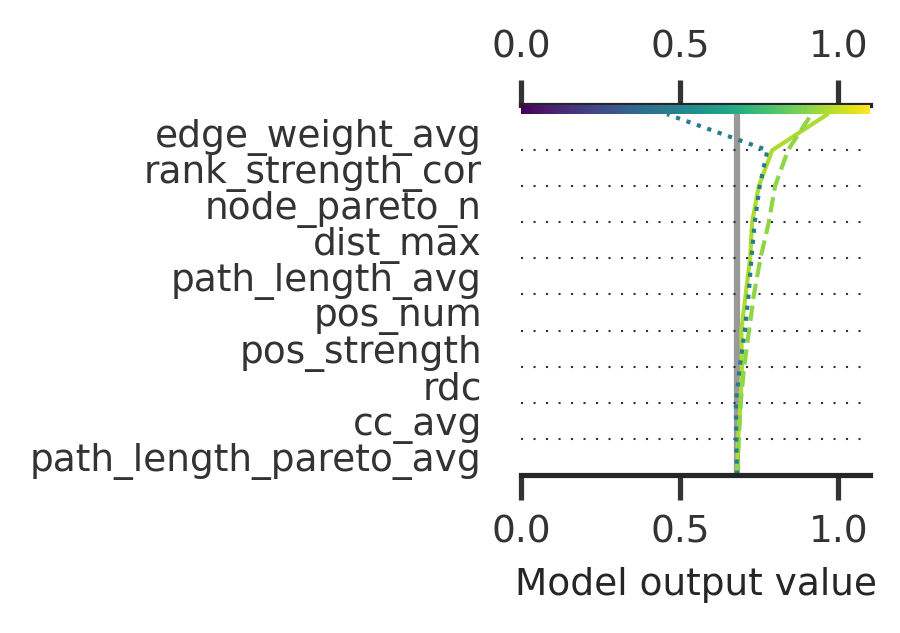}
        \caption{$\rho=0.0$, $m=3$, $k=1$}
        \label{fig:subfig10}
    \end{subfigure}
    \hfill
    \begin{subfigure}[t]{0.25\textwidth}
        \centering
        \includegraphics[width=\textwidth]{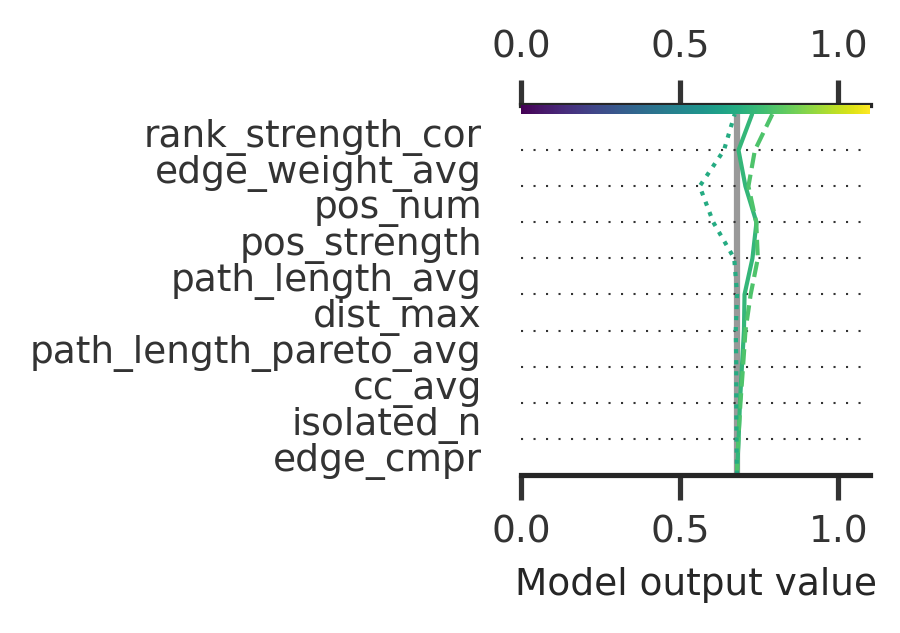}
        \caption{$\rho=0.0$, $m=3$, $k=2$}
        \label{fig:subfig11}
    \end{subfigure}
    \hfill
    \begin{subfigure}[t]{0.25\textwidth}
        \centering
        \includegraphics[width=\textwidth]{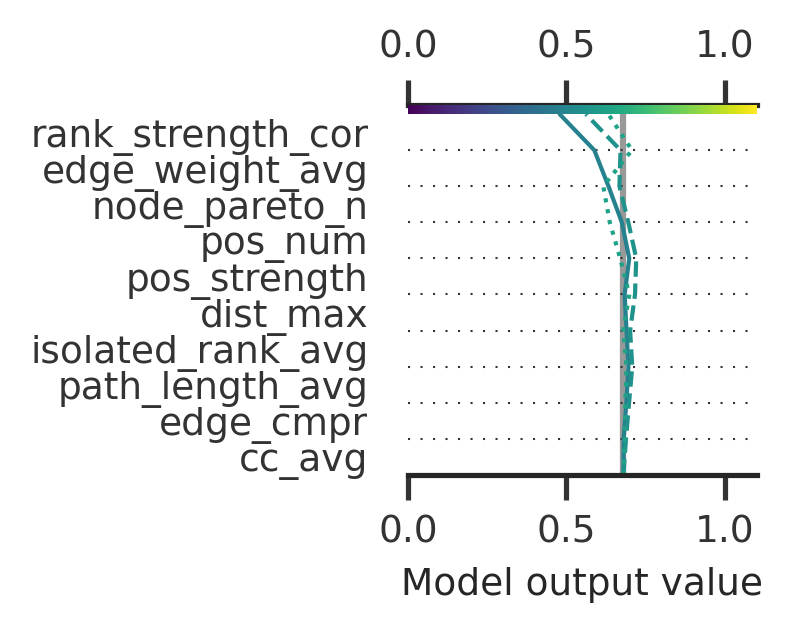}
        \caption{$\rho=0.0$, $m=3$, $k=4$}
        \label{fig:subfig12}
    \end{subfigure}
    
    \vspace{0.1cm} 
    
    \begin{subfigure}[t]{0.25\textwidth}
        \centering
        \includegraphics[width=\textwidth]{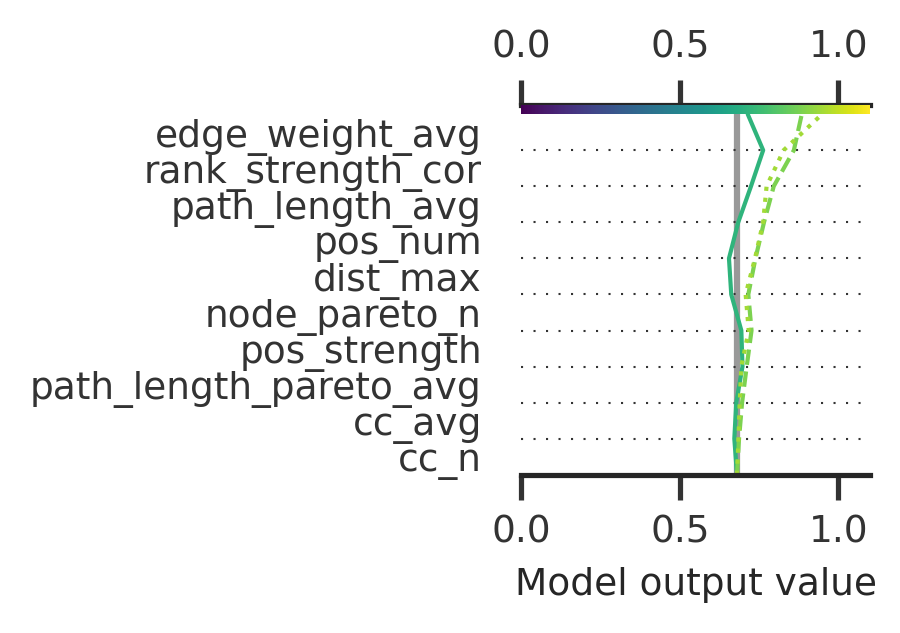}
        \caption{$\rho=0.4$, $m=2$, $k=1$}
        \label{fig:subfig13}
    \end{subfigure}
    \hfill
    \begin{subfigure}[t]{0.25\textwidth}
        \centering
        \includegraphics[width=\textwidth]{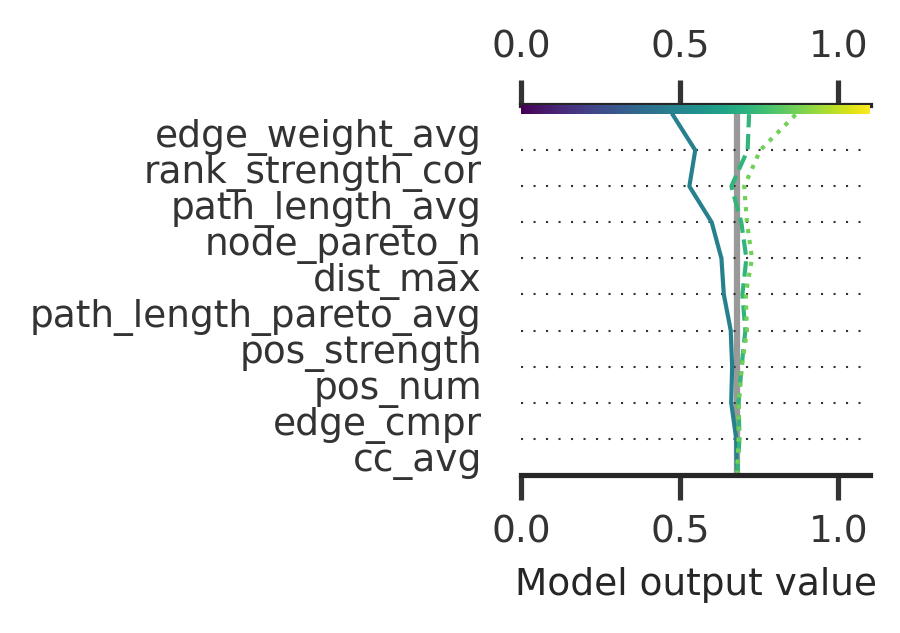}
        \caption{$\rho=0.4$, $m=2$, $k=2$}
        \label{fig:subfig14}
    \end{subfigure}
    \hfill
    \begin{subfigure}[t]{0.25\textwidth}
        \centering
        \includegraphics[width=\textwidth]{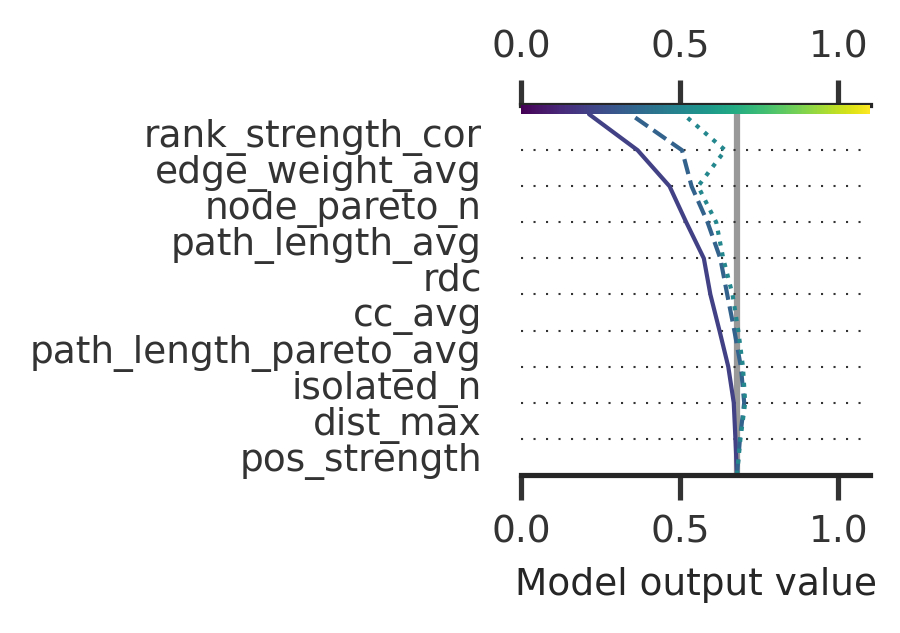}
        \caption{$\rho=0.4$, $m=2$, $k=4$}
        \label{fig:subfig15}
    \end{subfigure}
    
    \vspace{0.1cm} 
    
    \raisebox{23pt}{ 
    \begin{subfigure}[t]{0.25\textwidth}
        \centering
        \includegraphics[width=\textwidth]{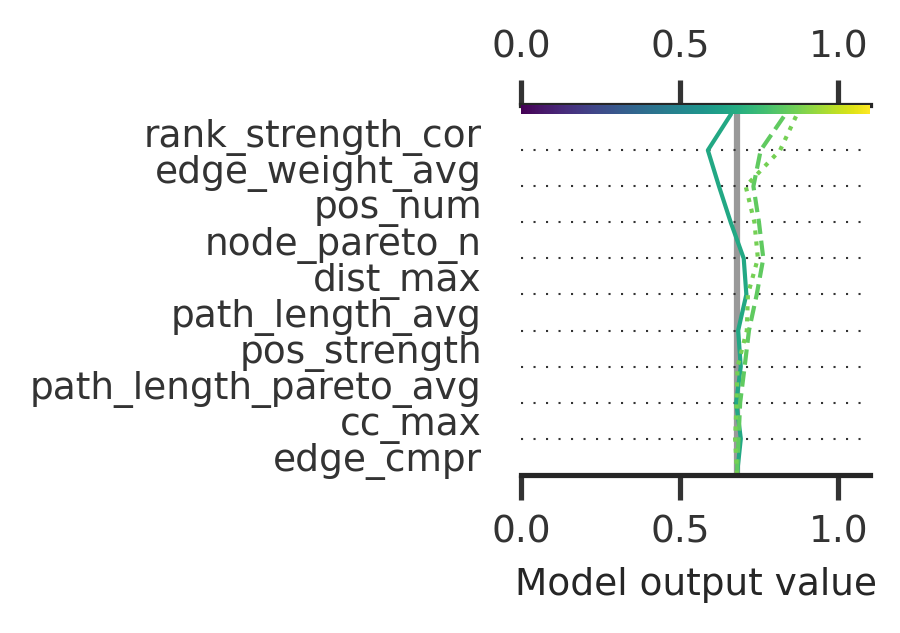}
        \caption{$\rho=0.4$, $m=3$, $k=1$}
        \label{fig:subfig16}
    \end{subfigure}}
    \hfill
    \begin{subfigure}[t]{0.27\textwidth}
        \centering
        \includegraphics[width=\textwidth]{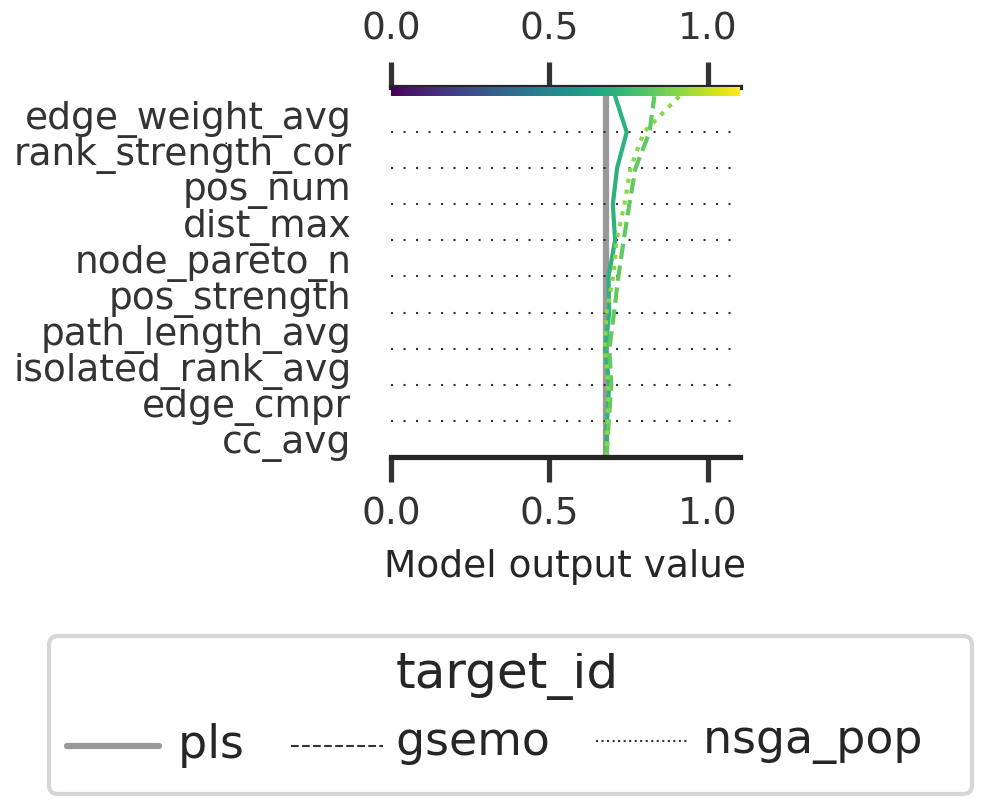} 
        \caption{$\rho=0.4$, $m=3$, $k=2$}
        \label{fig:subfig17}
    \end{subfigure}
    \hfill
    \raisebox{23pt}{ 
    \begin{subfigure}[t]{0.25\textwidth}
        \centering
        \includegraphics[width=\textwidth]{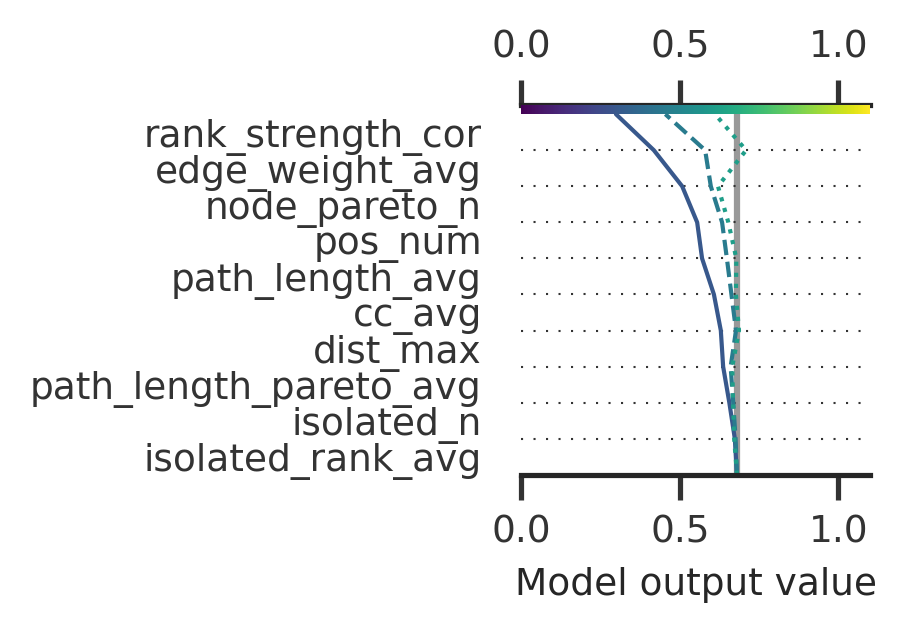}
        \caption{$\rho=0.4$, $m=3$, $k=4$}
        \label{fig:subfig18}
    \end{subfigure}}
    
    \caption{Key landscape features for $\rho$mnk landscapes are shown, with each plot representing a landscape and three lines for algorithm-feature interactions influencing performance: $\rho \in \{0, 0.4\}$, $m \in \{2,3\}$, $k \in \{1,2,4\}$. The x-axis shows the predicted $reso$ performance.}
    \Description{}
    \label{fig:post_hoc_second}
    
\end{figure*}

To formulate the footprints of the three algorithms, we create a contingency matrix that shows how the meta-representations associated with a single algorithm are distributed across the clusters. Figure~\ref{fig:footprint_reso} shows the algorithm footprints, with the rows representing the 14 clusters and the columns representing the $\rho$mnk problem instances generated with different benchmark parameters. The color coding reflects the predicted~\textsf{reso} performance. The sub-figures display the footprints for PLS (top), GSEMO (middle), and NSGA-II (bottom). By analyzing each footprint separately, we observe that PLS performs exceptionally well in a specific landscape with low objective correlation ($\rho = -0.4$) and a high number of objectives ($k = 3$). However, as both the objective correlation and the number of interactions increase ($\rho \in \{0.0, 0.4\}$, particularly for $k \in \{2, 3\}$), its performance drastically deteriorates. On the other hand, GSEMO demonstrates good performance in many problem instances and is less sensitive to changes in objective correlation. However, its performance significantly drops when the number of interactions is very high ($k = 4$). Finally, NSGA-II exhibits similar behavior to GSEMO, but seems to be also affected by an increase in the number of objectives, especially when the number of interactions is set to four or for low and medium objective correlations ($\rho \in \{0.0, 0.4\}$).

By comparing the three footprints, we can see to which cluster the same $\rho$mnk problem instance belongs for each of them (thus analyzing column-wise across sub-figures). Focusing on the first three $\rho$mnk problem instances, corresponding to $\rho = -0.4$, $m = 2$, and $k \in {1, 2, 4}$, we can observe that NSGA-II efficiently solves these landscapes except in the case of very high number of interactions ($k$=4). A similar pattern can be observed for GSEMO, although driven by a different important landscape feature combination, as the clusters differ between the two algorithms. In contrast, PLS performs worse than both NSGA-II and GSEMO, struggling even for moderate interactions ($k = 2$) and showing significant performance degradation when the number of interactions becomes very high ($k = 4$). Examining the next three columns, corresponding to $\rho = -0.4$, $m = 3$, and $k \in \{1, 2, 4\}$, we observe notable differences in algorithm performance. PLS excels in solving these landscapes, demonstrating highly effective performance across all cases. GSEMO shows moderate performance, handling these problem instances reasonably well but not matching the effectiveness of PLS. In contrast, NSGA-II performs poorly, struggling significantly in these landscapes. These results highlight the complementarity of the algorithms.

\begin{figure*}[t]
    \centering
    \begin{subfigure}[t]{0.2\textwidth}
        \centering
        \includegraphics[width=\textwidth]{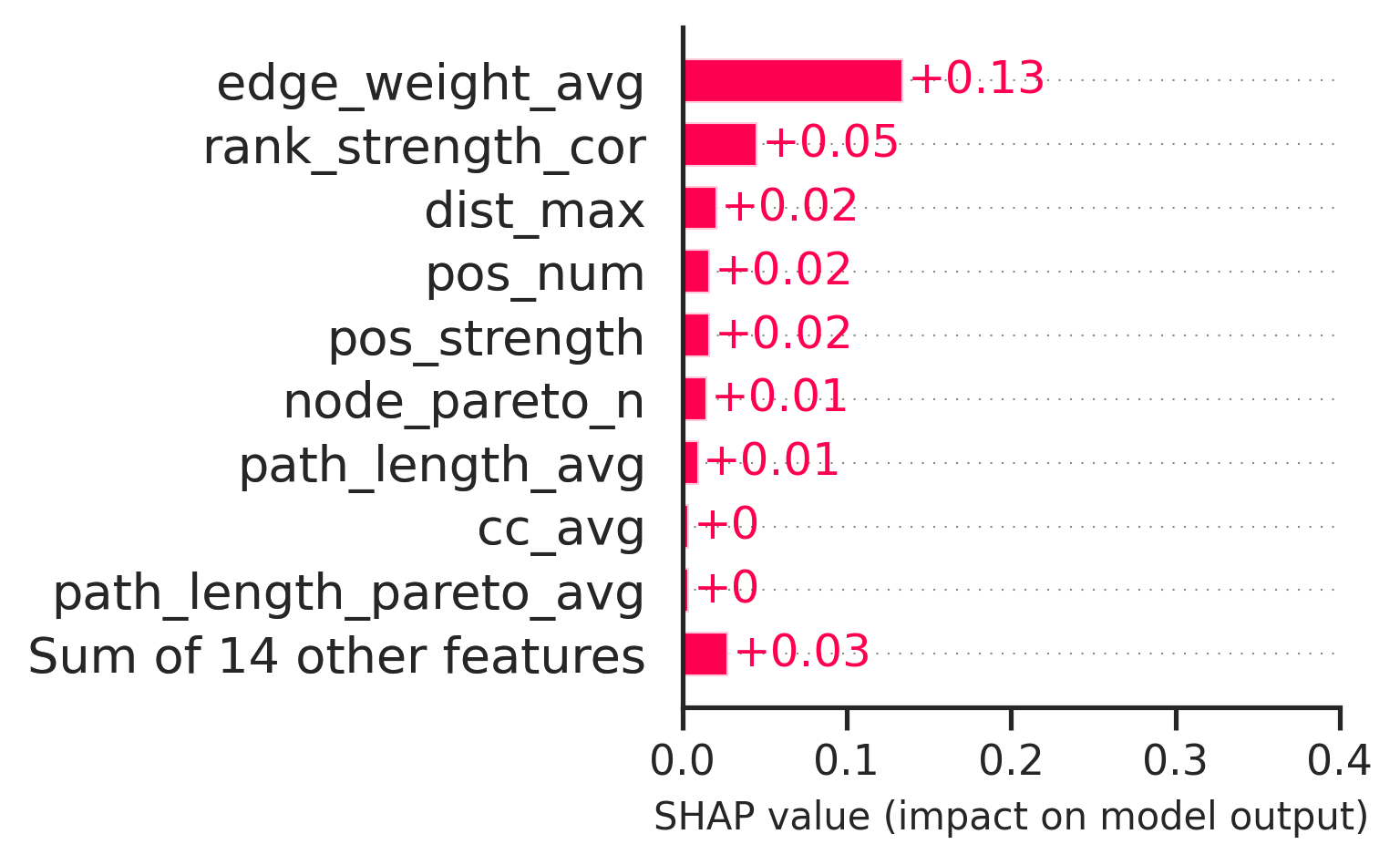}
        \caption{First cluster.}
        \label{fig:subfig111}
    \end{subfigure}
    \hfill
    \begin{subfigure}[t]{0.2\textwidth}
        \centering
        \includegraphics[width=\textwidth]{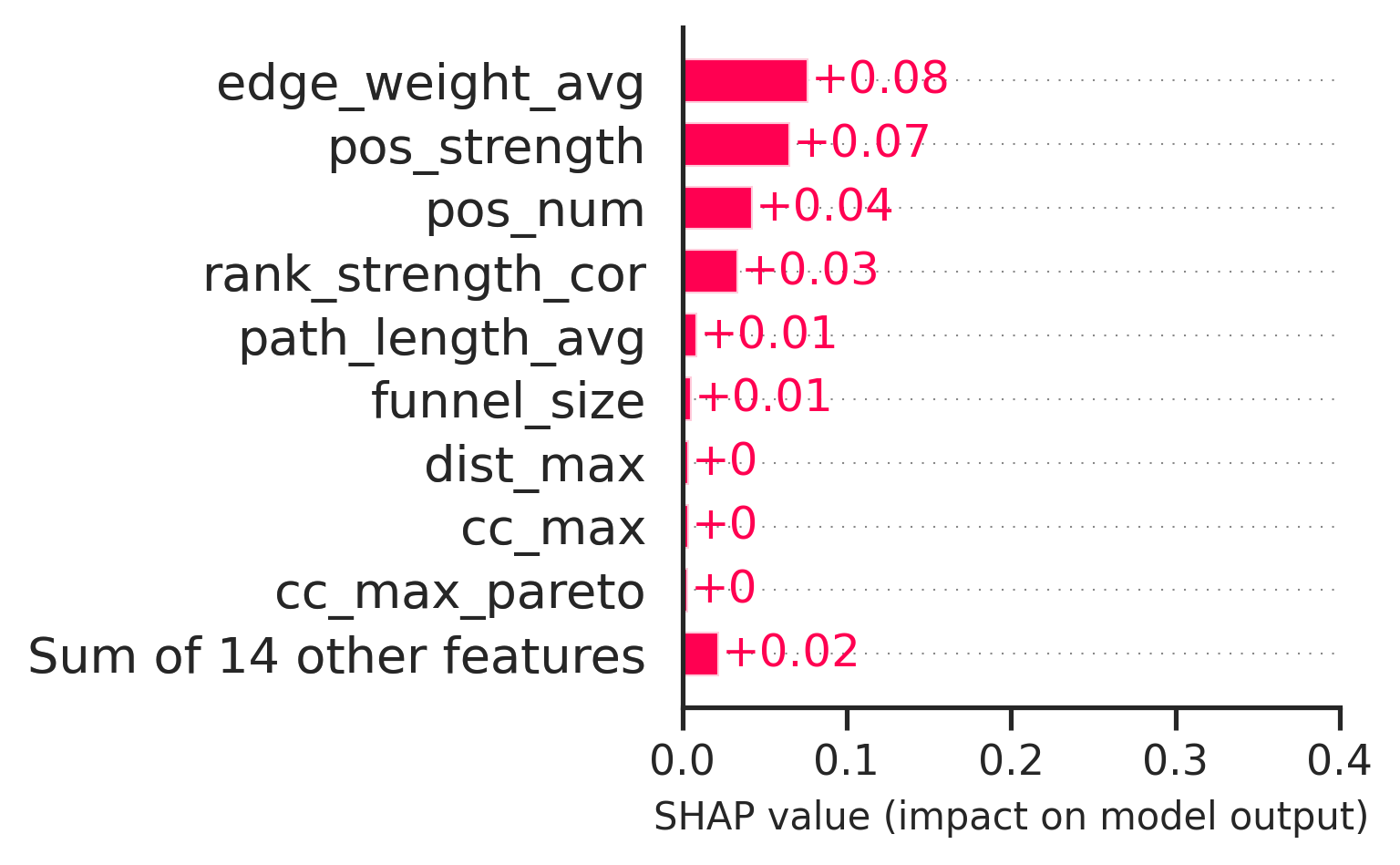}
        \caption{Seventh cluster.}
        \label{fig:subfig333}
    \end{subfigure}
    \hfill
    \begin{subfigure}[t]{0.2\textwidth}
        \centering
        \includegraphics[width=\textwidth]{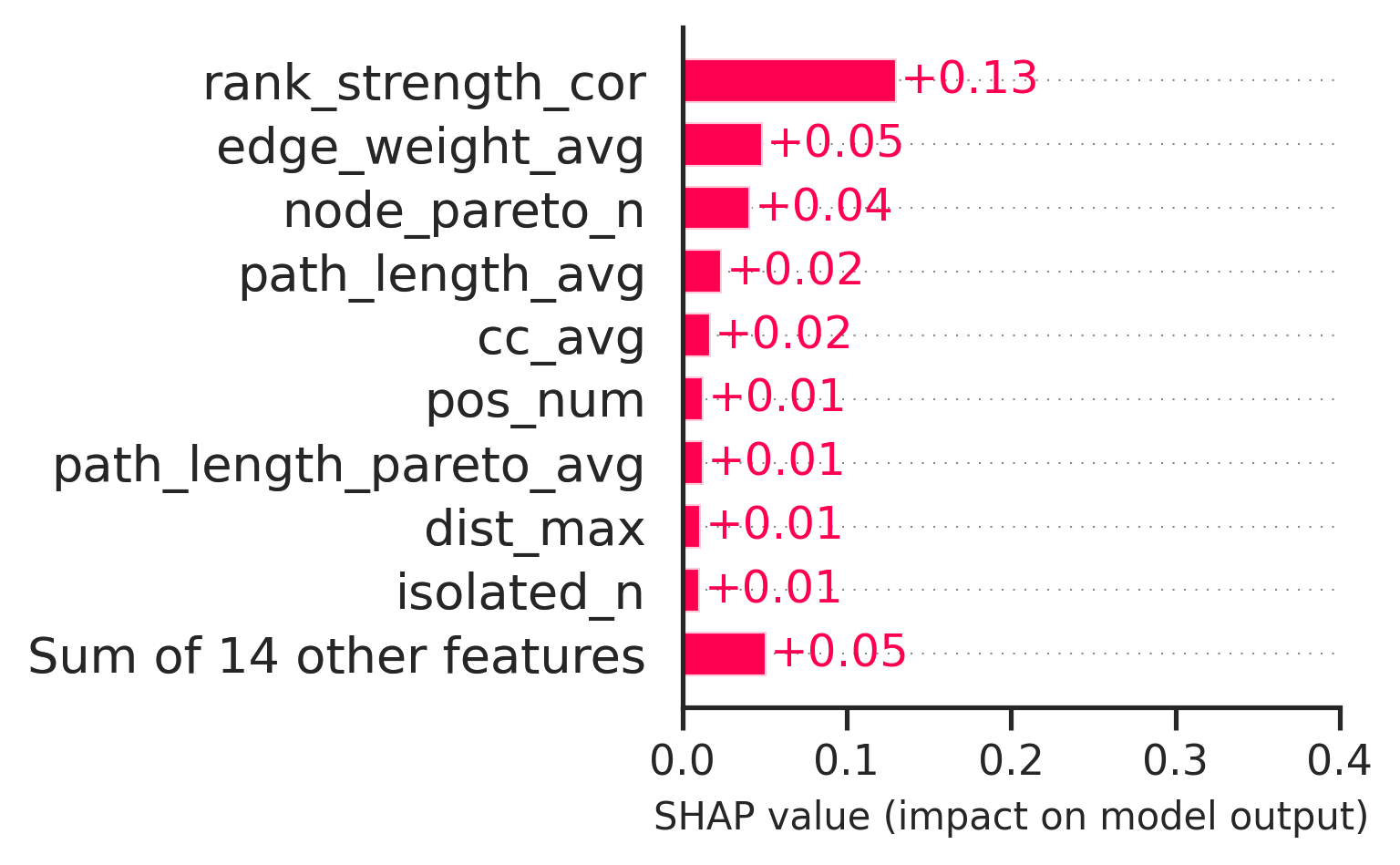}
        \caption{Thirteenth cluster.}
        \label{fig:subfig555}
    \end{subfigure}
    \hfill
    
    \vspace{0.1cm} 
    
     \begin{subfigure}[t]{0.2\textwidth}
        \centering
        \includegraphics[width=\textwidth]{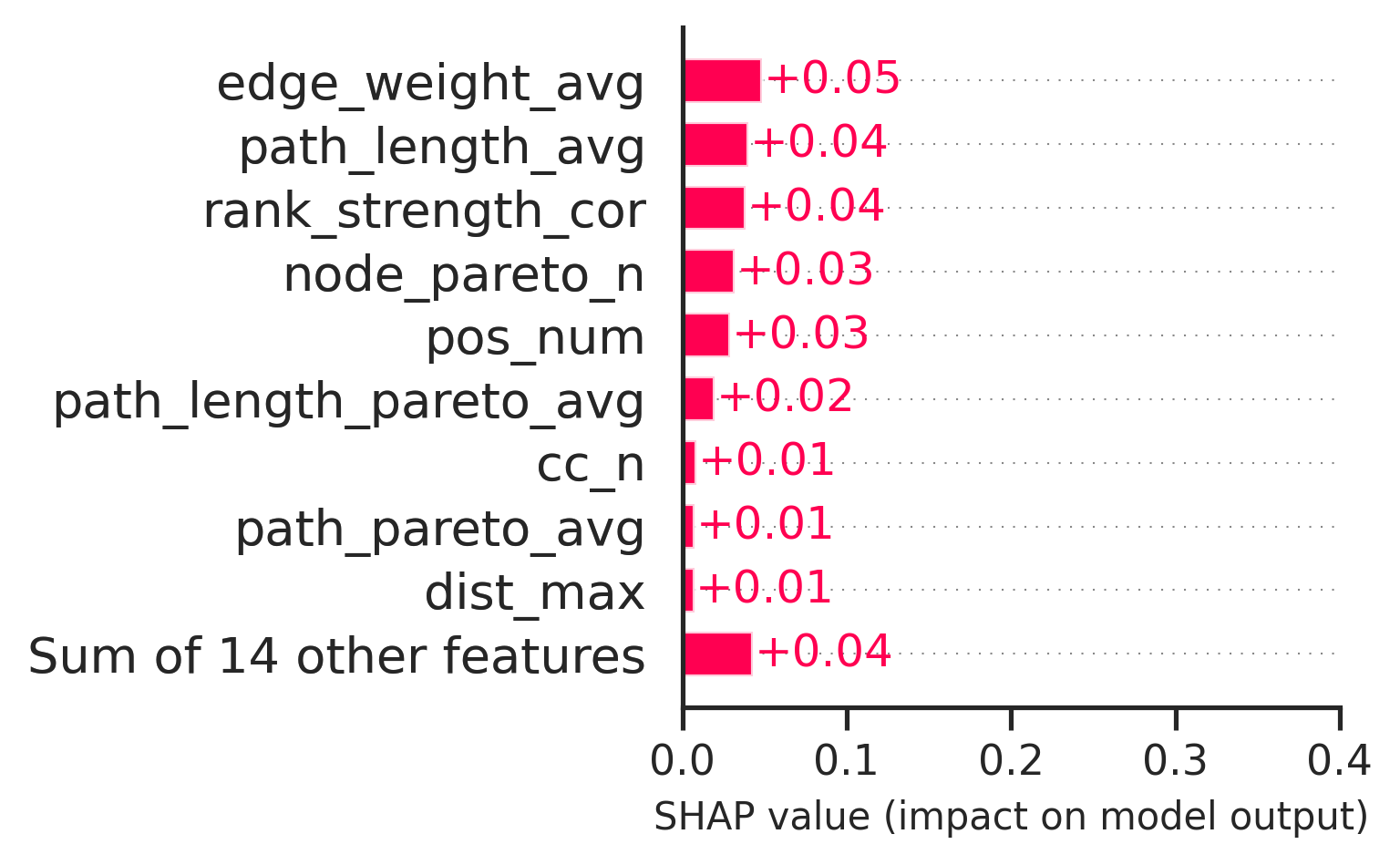}
        \caption{Second cluster.}
        \label{fig:subfig222}
    \end{subfigure}
    \hfill
    \begin{subfigure}[t]{0.2\textwidth}
        \centering
        \includegraphics[width=\textwidth]{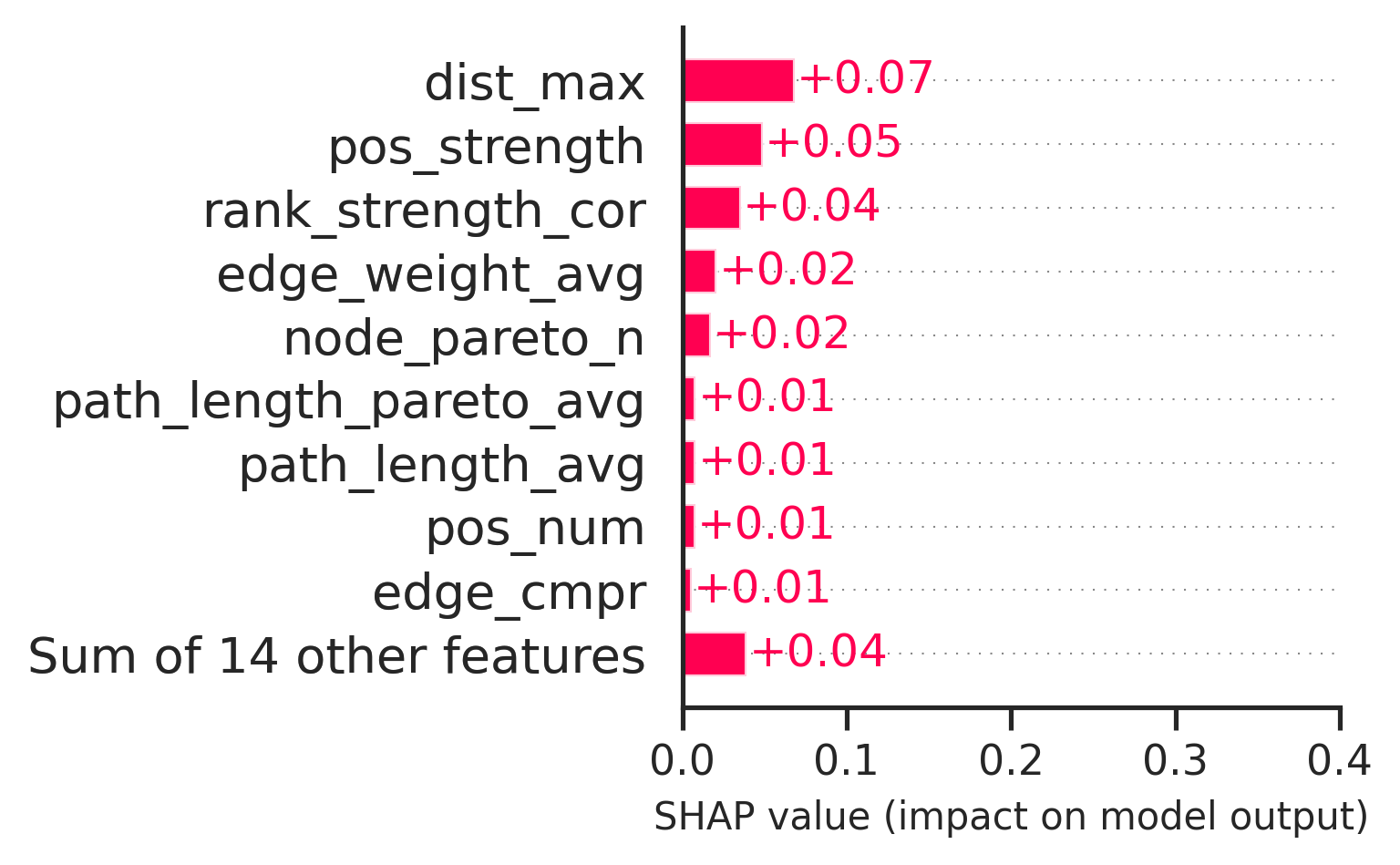}
        \caption{Eight cluster.}
        \label{fig:subfig444}
    \end{subfigure}
    \hfill
    \begin{subfigure}[t]{0.25\textwidth}
        \centering
        \includegraphics[width=\textwidth]{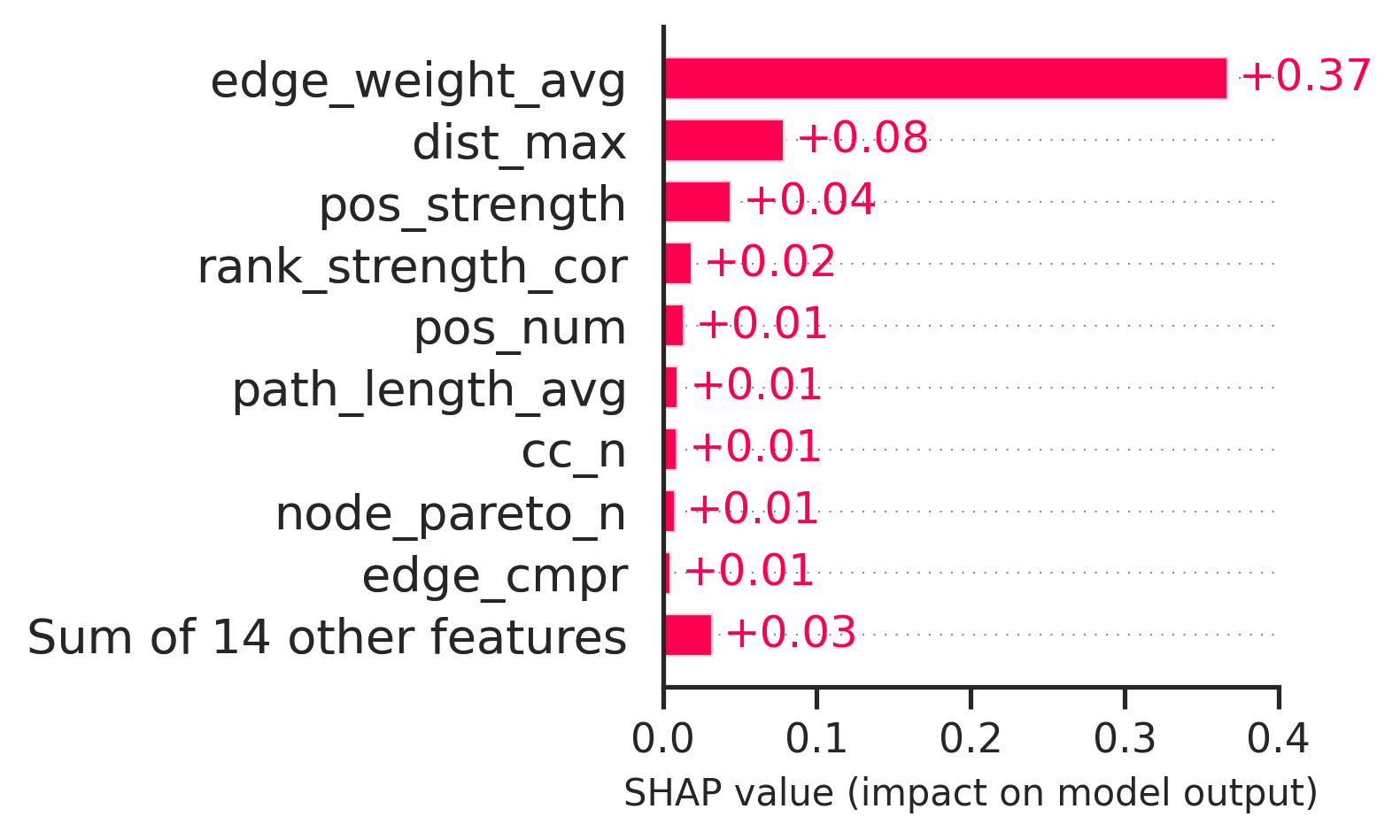}
        \caption{Fourteenth cluster.}
        \label{fig:subfig666}
    \end{subfigure}
    
    \caption{The key features driving \textsf{reso} performance are shown for each cluster (performance region). Analysis includes the top two best, two medium, and two worst-performing regions.}
    \Description{}
    \label{fig:global_features_reso_selected}
\end{figure*}

\vspace{-1em}
\subsubsection{Important landscape feature combinations across different $\rho$mnk problem instances.} Figures~\ref{fig:post_hoc_first} and~\ref{fig:post_hoc_second} showcase the most important landscape feature combination for each of the 18 benchmark parameter combinations. Each sub-figure displays a SHAP~\textit{decision plot}, a powerful visualization used to explain the predictions made by the MTR model. In these plots, the y-axis lists the landscape features ranked by their importance, with the most important features positioned at the top. The x-axis corresponds to the model's prediction for~\textsf{reso}, with the base value (the model’s prediction in the absence of any landscape feature information) marked as a vertical line. This base value serves as the reference point, indicating what a baseline model would predict when no landscape features are provided. Additionally, the plot includes segmented lines, where each segment corresponds to a landscape feature. The deviation of the line segment from the base value represents how much the feature influences the model's prediction. The three lines trace the path from the base value to the final model predictions for each algorithm separately, allowing for a clear comparison of feature importance across the three algorithms. The line color reflects predicted reso performance, ranging from dark blue (the worst) to yellow (the best). The line style distinguishes between the three algorithms with the solid line corresponding to PLS, the dashed to GSEMO, and the dotted line to NSGA-II, as indicated by the legend at the bottom of the plot.

Focusing on Figure~\ref{fig:subfig2} ($\rho=-0.4$, $m=3$, and $k=1$), we can see the combination of the landscape features that are causing the bad performance of NSGA-II, the medium performance of GSEMO, and the good performance of PLS. By examining the feature importance, we can conclude that some features, such as \textsf{rdc}, \textsf{cc\_n}, and \textsf{edge\_cmpr}, play a minimal role in determining the model’s predictions, as the line segments remain close to the base value. The other landscape features starting from~\textsf{pos\_num} have a much larger impact on the predictions. This means that they are the characteristics of this landscape that make it particularly challenging for NSGA-II but relatively easy for PLS to navigate, among which~\textsf{edge\_weight\_avg} holds the highest importance. We argue that smooth landscapes ($k=1$) generally have a single funnel, thus local search can effectively follow direct heavily visited (i.e. high edge weight) ``up-hill" pathways  leading to good solutions. For GSEMO the landscape features from~\textsf{node\_pareto\_n} to~\textsf{rank\_strength\_cor} contribute to predicting good ~\textsf{reso} performance. However,~\textsf{dist\_max} and~\textsf{edge\_weight\_avg} prevent GSEMO from achieving near-optimal performance (reso close to 1). For the PLS algorithm, similar patterns are observed, but~\textsf{dist\_max} and ~\textsf{edge\_weight\_avg} become advantageous landscape features, enabling near-perfect performance. Again, this is probably due to the most effective, greedy search that PLS can achieve in relatively smooth landscapes.

Next, from the last column of sub-figures ($k=4$) the combination of the landscape features that are causing bad performance for all the algorithms become evident. Across all the plots, we consistently observe that the features \textsf{rank\_strength\_cor}, \textsf{edge\_weight\_avg}, and \textsf{node\_pareto\_n} emerge as the most important landscape features. These features influence all three algorithms, albeit to varying extents. This suggests that the PLS algorithm struggles to handle landscapes characterized by these features, whereas NSGA-II is able to achieve better performance to some extent. Due to space constraints, we will not present explanations for all benchmark parameter combinations. However, the figures provide insights into the importance of various landscape features, tailored to specific \rmnk s and algorithms. 

In this study, we examined landscape features both in isolation and in combination, revealing that the two new proposed funnel features from this study, such as~\textsf{pos\_num} and~\textsf{pos\_strength}, consistently appear as important across nearly all landscapes, while~\textsf{rdc} is crucial for several. Among the landscapes where~\textsf{rdc} plays a role, its most significant contribution occurs for $\rho=0.4$, $m=2$, and $k=4$. This feature, alongside others, makes this particular landscape challenging for all algorithms. It is important to note that the landscape features presented here represent a selection of those that contribute to the best predictive MTR model. For instance, other funnel features correlate with~\textsf{path\_length\_avg} and~\textsf{path\_pareto\_avg}, which are part of the selected set. In the future, if a specific feature portfolio needs to be evaluated, the corresponding footprints can be easily calculated.
\vspace{-1em}
\subsubsection{Important landscape feature combinations across clusters.} 
Each cluster contains problem instances with similar~\textsf{reso} performance (good, medium, or bad), achieved by different algorithms, due to similar reasons (similar most important feature combination). However, when two clusters contain problem instances with similar \textsf{reso} performance, the underlying reasons for achieving that performance differ (distinct most important feature combinations between the clusters). Figure~\ref{fig:global_features_reso_selected} presents the most important feature combination for each cluster. The analysis focuses on a subset of clusters, including two clusters with good algorithm performance (the first and second clusters), two with medium performance (the seventh and eighth clusters), and two with bad performance (the thirteenth and fourteenth clusters). When examining cluster pairs column-wise—clusters share similar \textsf{reso} performance -- the most important feature combination is distinct. For example, the importance of the~\textsf{edge\_weight\_avg} feature is significantly more pronounced in the first cluster compared to the second. Further, when comparing clusters row-wise-clusters with different \textsf{reso} performance (ranging from good to bad) -- the most important feature combinations vary substantially. Features such as~\textsf{pos\_strenght},~\textsf{pos\_num} and~\textsf{dist\_max} become important. Note that we show absolute feature importance, so the same features may matter in both good and bad \textsf{reso} clusters, though their influence (direction) differs as explained earlier.

\vspace{-1em}
\section{Discussion}
\label{sec:discussion}
We analyzed the~\textsf{hv} metric, but the results are omitted due to space (available in our repository). As algorithms show little complementarity on this metric, landscape features have a similar influence, though interaction patterns vary specific to the landscapes. For the~\textsf{hv}, the clustering resulted in 14 regions. Key features also differ from those for the~\textsf{reso} metric. The features that appear among the most important features are~\textit{cc\_avg},~\textit{sink\_num},~\textit{sink\_strenght},\\~\textit{pos\_strenght},~\textit{dist\_pareto\_avg},~\textit{dist\_max},~\textit{rank\_strength\_cor},\\~\textit{node\_pareto\_n}, and~\textit{assort\_degree}. Algorithm footprints are based on MTR model predictions, which, despite minor errors, closely approximate ground-truth \textsf{reso} performance. While our method uses an MTR predictive model, increasing the number of targets poses challenges due to the curse of dimensionality, a known issue in ML. Addressing this, deep learning techniques and larger benchmark sets could enhance MTR models to handle more algorithms. The applicability of the approach can be extended to larger-scale multi-objective combinatorial problems or different combinatorial structures, provided that relevant data can be collected. 

\vspace{-1em}
\section{Conclusion}
\label{sec:conclusions}
 By features from the C-PLOS-net model in combination with two newly introduced funnel features, we analyzed benchmark \rmnk-landscapes with varying ruggedness and objective correlations. The evaluation of PLS, GSEMO, and NSGA-II using resolution metric revealed key feature combinations that influence algorithm performance. Additionally, the analysis highlights challenges specific to certain landscapes, offering deeper insights into feature importance and its implications for algorithm design and benchmarking in multi-objective optimization. The findings highlight the number of interactions as the most influential benchmarking parameter shaping problem difficulty, providing valuable insights for algorithm selection and optimization strategies. Furthermore, the study revealed that combining two sets of features is essential, as their combination emerges as the most significant for specific landscapes when paired with different algorithms. This indirectly provides empirical evidence that these feature sets offer complementary information about the characteristics of the landscapes.



\begin{acks}
We acknowledge the support of the Slovenian Research and Innovation Agency through program grant P2-0098, and project grants No.J2-4460 and No. GC-0001, and young researcher grant No. PR-12897 to AN. This work is also funded by the European Union under Grant Agreement 101187010 (HE ERA Chair AutoLearn-SI).
\end{acks}

\appendix

\end{document}